\documentclass{article}
\usepackage{microtype}
\usepackage{subcaption}
\usepackage{booktabs} %

\usepackage{hyperref}
\usepackage{amsthm,amssymb}
\usepackage[toc,page]{appendix}
\usepackage{graphicx}

\usepackage{amsmath,amsfonts,bm}

\def\eqref#1{equation~\ref{#1}}

\def\1{\bm{1}}

\DeclareMathAlphabet{\mathsfit}{\encodingdefault}{\sfdefault}{m}{sl}
\SetMathAlphabet{\mathsfit}{bold}{\encodingdefault}{\sfdefault}{bx}{n}

\usepackage{color,xcolor}
\usepackage{bm}
\usepackage{amsmath}
\usepackage{mathtools}
\usepackage{adjustbox}
\usepackage{multirow}
\usepackage{algorithm}

\usepackage{amsmath,amsfonts,bm}

\def\eqref#1{equation~\ref{#1}}

\def\1{\bm{1}}

\DeclareMathAlphabet{\mathsfit}{\encodingdefault}{\sfdefault}{m}{sl}
\SetMathAlphabet{\mathsfit}{bold}{\encodingdefault}{\sfdefault}{bx}{n}

\newcommand{\cB}{\mathcal{B}}
\newcommand{\cS}{\mathcal{S}}
\newcommand{\cD}{\mathcal{D}}

\newcommand{\cW}{\mathcal{W}}

\newcommand{\cX}{\mathcal{X}}
\newcommand{\cY}{\mathcal{Y}}

\newcommand{\cT}{\mathcal{T}}

\newcommand{\bbR}{\mathbb{R}}
\newcommand{\bbE}{\mathbb{E}}

\usepackage[accepted]{icml2021}
\icmltitlerunning{Voice2Series: Reprogramming Acoustic Models for Time Series Classification}

\begin{document}

\twocolumn[
\icmltitle{Voice2Series: Reprogramming Acoustic Models for Time Series Classification }

\icmlsetsymbol{equal}{*}

\begin{icmlauthorlist}
\icmlauthor{Chao-Han Huck Yang}{a}
\icmlauthor{Yun-Yun Tsai}{b}
\icmlauthor{Pin-Yu Chen}{c}
\end{icmlauthorlist}

\icmlaffiliation{a}{Georgia Institute of Technology}
\icmlaffiliation{b}{Columbia University}
\icmlaffiliation{c}{IBM Research}

\icmlcorrespondingauthor{Chao-Han Huck Yang}{huckiyang@gatech.edu}
\icmlcorrespondingauthor{Pin-Yu Chen}{pin-yu.chen@ibm.com}
\icmlkeywords{Machine Learning, ICML}

\vskip 0.3in
]
\printAffiliationsAndNotice{} %

\begin{abstract}

Learning to classify time series with limited data is a practical yet challenging  problem.
Current methods are primarily based on hand-designed feature extraction rules or domain-specific data augmentation. Motivated by the advances in deep speech processing models and the fact that
voice data are univariate temporal signals,
in this paper, we propose \textit{Voice2Series} (V2S), 
a novel end-to-end approach that reprograms acoustic models for time series classification, through input transformation learning and output label mapping.
Leveraging the representation learning power of a large-scale pre-trained speech processing model, on 30 different time series tasks we show that V2S performs competitive results on \textbf{19} time series classification tasks. 
We further provide a theoretical justification of V2S by proving its population risk is upper bounded by the source risk and a Wasserstein distance accounting for feature alignment via reprogramming. Our results offer new and effective means to time series classification.
Our code of is available at \url{https://github.com/huckiyang/Voice2Series-Reprogramming}.
\end{abstract}

\section{Introduction}
\label{sec:1}

Machine learning for time series data has rich applications in a variety of domains, ranging from medical diagnosis (e.g.,
physiological signals such as electrocardiogram (ECG)~\cite{kampouraki2008heartbeat}), 
finance/weather forecasting, to industrial measurements (e.g., sensors and Internet of Things (IoT)). 
It is worth noting that one common practical challenge that prevents time series learning tasks from using modern large-scale deep learning models is data scarcity. While many efforts~\cite{fawaz2018transfer, ye2018novel, kashiparekh2019convtimenet} have been made to advance transfer learning and model adaptation for time series classification, a principled approach is lacking and its performance may not be comparable to conventional statistical learning benchmarks~\cite{langkvist2014review}.

\begin{figure}[t]
\centering
\includegraphics[width=0.47\textwidth]{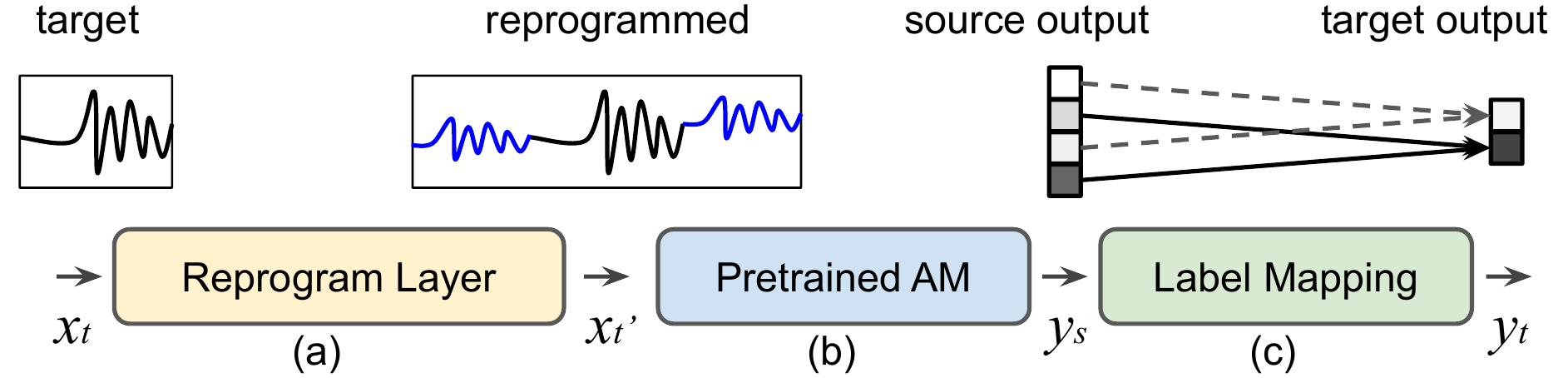}
\caption{Schematic illustration of the proposed Voice2Series (V2S) framework: (a) trainable reprogram layer; (b) pre-trained acoustic model (AM); (c) source-target label mapping function.  }
\label{fig:1:sys}
\vspace{-2mm}
\end{figure}

To bridge this gap, we propose a novel approach, named \textbf{voice to series (V2S)}, for time series classification by \textit{reprogramming} a pre-trained acoustic model (AM), such as a spoken-terms recognition model. Unlike general time series tasks, modern AMs are trained on massive human voice datasets and are considered as a mature technology widely deployed in intelligent electronic devices.
The rationale of V2S lies in the fact that voice data can be viewed as univariate temporal signals, and therefore a well-trained AM is likely to be reprogrammed as a powerful feature extractor for solving time series classification tasks. Figure \ref{fig:1:sys} shows a schematic illustration of the proposed V2S framework, including (a) a trainable reprogram layer, (b) a pre-trained AM, and (c) a specified label mapping function between source (human voice) and target (time series) labels.

Model reprogramming was firstly introduced in 
\cite{elsayed2018adversarial}. The authors show that one can learn a universal input transformation function to reprogram a pre-trained ImageNet model (without changing the model weights) for solving MNIST/CIFAR-10 image classification and simple vision-based counting tasks with high accuracy. It can be viewed as an efficient approach for transfer learning with limited data, and it has achieved state-of-the-art (SOTA) results on biomedical image classification tasks \cite{tsai2020transfer}. However, despite the empirical success, little is known on how and why reprogramming can be successful.

Different from existing works, this paper aims to address the following three fundamental questions: (i) Can acoustic models be reprogrammed for time series classification? (ii) Can V2S outperform SOTA time-series classification results? (iii) Is there any theoretical justification on why reprogramming works?

Our main contributions in this paper provide affirmative answers to the aforementioned fundamental questions, which are summarized as follows.
\begin{enumerate}
    \item We propose V2S, a novel and unified approach to reprogram large-scale pre-trained acoustic models for different time series classification tasks. To the best of our knowledge, V2S is the first framework that
    enables reprogramming for time series tasks. 
    \item Tested on a standard UCR time series classification benchmark \cite{dau2019ucr}, V2S performs competitively on \textbf{19} out of 30 datasets, suggesting that V2S is a potentially effective approach for time series classification.
    \item In Section \ref{sec_risk}, we develop a theoretical risk analysis to characterize the performance of reprogramming on the target task via source risk and representation alignment loss. In Section \ref{sec_exp}, we also show how our theoretical results can be used to assess the performance of reprogramming. Moreover, we provide interpretation on V2S through auditory neural saliency map and embedding visualization.  
\end{enumerate}

\section{Related Works}
\label{sec:2}

\subsection{Time Series Classification}
Learning to classify time series is a standard research topic in 
machine learning and signal processing. A major research branch uses designed features followed by conventional classifiers, such as
digital filter design in together with support vector machine (SVM)~\cite{kampouraki2008heartbeat}, decision-trees~\cite{geurts2001pattern}, or kernel based methods~\cite{zhang2010time, lines2012shapelet}. 
Recently, deep learning models have been utilized in time series~\cite{fawaz2019deep} and demonstrated improved performance.
The methods range from a completely end-to-end classifier~\cite{zhang2010time, wang2017time} to a mixture model~\cite{hong2019mina} that combines feature engineering  and deep learning. Notably, feature engineering methods~\cite{wang2017time} 
can still attain competitive results on some time series classification tasks, especially when the number of training data is small.

\subsection{Model Reprogramming}
Although in the original paper~\cite{elsayed2018adversarial} model reprogramming was phrased as ``adversarial'' reprogramming, it is not limited to the adversarial setting, nor does it involve any adversarial training. \citet{elsayed2018adversarial} showed that the pre-trained ImageNet models can be reprogrammed for classifying other image datasets and for solving vision-based counting tasks. \citet{tsai2020transfer} demonstrated the advantage of reprogramming on label-limited data such as biomedical image classification, and used zeroth order optimization \cite{liu2020primer} to enable reprogramming of black-box machine learning systems. 

Beyond image data, model reprogramming has been used in natural language processing (NLP) \cite{neekhara2019adversarial,hambardzumyan2020warp}, such as machine translation and sentiment classification. \citet{vinod2020reprogramming} further showed that NLP models can be reprogrammed for molecule learning tasks in biochemistry.

For reprogramming with time series, it remains unclear what source domain and pre-trained models to be used. Current research works on reprogramming do not extend to time series because
image and text data have fundamentally different feature characteristics than time series. One of our major contributions is the proposal of reprogramming pre-trained acoustic models on abundant human voice data for time series classification. To our best knowledge, our V2S is the first framework on  reprogramming for time series.

\subsection{Deep Acoustic Modeling}
Recent deep learning models have shown impressive results on predicting label(s) from acoustic information. The central idea is to use a large number of spectral features (e.g., Mel-spectrogram or log-power spectrum) as training inputs to capture the important features. Some of the latent features learned from AM are interpretable based on physiological auditory experiments (e.g., cortex responses~\cite{kaya2017modelling}) or neural saliency methods. Several efforts have been put into the design of a large and deep neural network to extract
features from human voice datasets.
Among them,
residual neural network (ResNet)~\cite{he2016deep, saon2017english} and VGGish~\cite{hershey2017cnn} models are popular backbones for AM tasks, such as spoken-term recognition~\cite{yang2020decentralizing, de2018neural} and speech enhancement~\cite{xu2014regression, yang2020characterizing}.
It is worth noting that standard transfer learning via parameter finetuning is not ideal for time series tasks with limited data, as acoustic data and time-series data are often quite diverse in scale. Our V2S addresses this issue via learning an input transformation function while fixing the pre-trained AM.

\begin{table*}[t]
\centering
\caption{Mathematical notation for reprogramming }
\label{tab:notation}
\begin{tabular}{@{}l|l@{}}
\toprule
Symbol                                                                                                    & Meaning                                                                   \\ \midrule
$\cS$ / $\cT$                                                                                             & source/target domain                                                    \\
$\cX_{\cS}$ / $\cX_{\cT}$                                                                                & the space of source/target data samples                                 \\
$\cY_{\cS}$ / $\cY_{\cT}$                                                                                 & the space of source/target data labels                                  \\
$\cD_{\cS}$ $\subseteq \cX_{\cS} \times \cY_{\cS} $ / $\cD_{\cT}$ $\subseteq \cX_{\cT} \times \cY_{\cT} $ & source/target data distribution                                         \\
$(x,y) \sim \cD$                                                                                          & data sample $x$ and one-hot coded label $y$ drawn from $\cD$              \\
$K$                                                                                                       & number of source labels                                                   \\
$f_\cS: \bbR^d \mapsto [0,1]^K $                                                                          & pre-trained  $K$-way source classification model                          \\
$\eta: \bbR^K \mapsto [0,1]^K$                                                                            & softmax function in neural network, and $\sum_{k=1}^K [\eta(\cdot)]_k =1$ \\
$z(\cdot) \in \bbR^{K}$                                                                                   & logit (pre-softmax) representation, and $f(x)=\eta(z(x))$                 \\
$\ell(x,y) \triangleq \|f(x)-y\|_2$                                                                                 & risk function of $(x,y)$ based on classifier $f$                          \\
$\bbE_{\cD} [\ell (x,y)] \triangleq \bbE_{(x,y) \ \sim \cD} [\ell (x,y)] = \bbE_{\cD} \|f(x)-y\|_2$               & population risk based on classifier $f$                                   \\ 
$\delta, \theta$             & additive input transformation on target data, parameterized by $\theta$                                 \\
\bottomrule
\end{tabular}
\end{table*}

\section{Voice2Series (V2S)}

\subsection{Mathematical Notation}

Table \ref{tab:notation} summarizes the major mathematical notations used in this paper for V2S reprogramming. Throughout this paper, we will denote a $K$-way acoustic classification model pre-trained on voice data as a \textit{source} model, and use the term \textit{target} data to denote the univariate time-series data to be reprogrammed. The notation $P$ is reserved for denoting a probability function. The remaining notations will be introduced when applicable.

\subsection{V2S Reprogramming on Data Inputs}
Here we formulate the problem of V2S reprogramming on data inputs.
Let $x_{t} \in \cX_{\cT} \subseteq \bbR^{d_{\cT}}$ denote a
univariate time series input from the target domain with $d_{\cT}$ temporal features. 
Our V2S aims to find a trainable input transformation function $\mathcal{H}$ that is universal to all target data inputs, which serves the purpose of reprogramming  $x_{t}$ into the source data space $\cX_{\cS} \subseteq \bbR^{d_{\cS}}$, where $d_{\cT} < d_{\cS}$. 
Specifically, the reprogrammed sample $x_t'$ is formulated as:
\begin{equation}
x_t' = \mathcal{H}(x_{t};\theta) := \text{Pad}(x_{t}) + \underbrace{M\odot \theta}_{\triangleq~\delta}
\label{eq:1:mask}
\end{equation}
where $\text{Pad}(x_{t})$ is a zero padding function that outputs a zero-padded time series of dimension $d_{\cS}$. The location of the segment $x_{t}$ to be placed in $x_t'$ is a design parameter and we defer the discussion to Section \ref{subsec_implement}.  The term
$M \in \{0,1\}^{d_{\cS}}$ is a binary mask that indicates the location of $x_{t}$ in its zero-padded input $\text{Pad}(x_{t})$, where the $i$-th entry of $M$ is 0 if $x_{t}$ is present (indicating the entry is non-reprogrammable), and it is $1$ otherwise (indicating the entry is not occupied and thus reprogrammable). The $\odot$ operator denotes element-wise product. Finally, $\theta \in \bbR^{d_{\cS}}$ is a set of trainable parameters for aligning source and target domain data distributions. One can consider a more complex function $W(\theta)$ in our reprogramming function. But in practice we do not observe notable gains when compared to the simple function $\theta$. In what follows, we will use the term $\delta \triangleq M\odot \theta$ to denote the trainable additive input transformation for V2S reprogramming. Moreover, for ease of representation we will omit the padding notation and simply use $x_t+ \delta$ to denote the reprogrammed target data, by treating the ``$+$'' operation as a zero-padded broadcasting function.

\subsection{V2S Reprogramming on Acoustic Models (AMs)}

We select a pre-trained deep acoustic classification model as the source model ($f_{\cS}$)  for model reprogramming. We assume the source model has softmax as the final layer and outputs nonnegative confidence score (prediction probability) for each source label.
With the transformed data inputs $\mathcal{H}(x_{t};\theta)$ described in (\ref{eq:1:mask}), one can obtain the class prediction of the source model $f_{\cS}$ on an reprogrammed target data sample $x_{t}$, denoted by 
\begin{equation}
P(y_s|f_{\cS}(\mathcal{H}(x_{t};\theta))), ~\text{for~all}~y_s \in \cY_{\cS}
\end{equation}

Next, as illustrated in Figure \ref{fig:1:sys}, we assign a (many-to-one) label mapping function $h$ to map source labels to target labels. For a target label $y_t \in \cY_{\cT}$, its class prediction will be the averaged class predictions over the set of source labels assigned to it. We use the term $P(h(\cY_{\cS})|f_{\cS}(\mathcal{H}(x_{t};\theta)))$ to denote the prediction probability of the target task on the associated ground-truth target label $y_t = h(\cY_{\cS})$. Finally, we learn the optimal parameters $\theta^*$ for data input reprogramming by optimizing the following objective:
\begin{align}
\label{eq:argmin}
& \theta^*  = \arg \min_{\theta} \underbrace{ -\log P(h(\cY_{\cS})|f_{\cS}(\mathcal{H}(x_{t};\theta))}_{\text{V2S~loss}~\triangleq~L};\\
& \text{where}~~~h\left(\cY_{\cS}\right) = y_{t} \nonumber
\end{align}
The optimization will be implemented by minimizing the empirical loss (V2S loss $L$) evaluated on all target-domain training data pairs $\{x_t,y_t\}$ for solving $\theta^*$.

In practice, we find that many-to-one label mapping can improve the reprogramming accuracy when compared to one-to-one label mapping, similar to the findings in \cite{tsai2020transfer}.
Below we make a concrete example on how many-to-one label mapping is used for V2S reprogramming. Consider the case of reprogramming spoken-term AM for ECG classification. One can choose to map multiple (but non-overlapping) classes from the source task (e.g., 'yes', 'no', 'up', 'down' in AM classes) to every class from the target task (e.g., 'Normal' or 'Ischemia' in ECG classes), leading to a specified mapping function $h$.
Let $\cB \subset \cY_{\cS}$ denote the set of source labels mapping to the target label $y_t \in \cY_{\cT}$. Then, the class prediction of $y_t $ based on V2S reprogramming is the aggregated prediction over the assigned source labels, which is defined  as 
\begin{align}
  \label{eq_prob}
  P(y_t|f_{\cS}(\mathcal{H}(x_{t};\theta)) = \frac{1}{|\cB|} \sum_{y_s \in \cB}  P(y_s|f_{\cS}(\mathcal{H}(x_{t};\theta))
\end{align}
where $|\cB|$ denotes the number of labels in $\cB$. In our implementation we use random (but non-overlapping) many-to-one mapping between source and target labels. Each target label is assigned with the same number of source labels.

\subsection{V2S Algorithm}

Algorithm \ref{V2S_training} summarizes the training procedure of our proposed V2S reprogramming algorithm. The algorithm uses the ADAM optimizer \cite{kingma2014adam} to find the optimal reprogramming parameters $\theta^*$ that minimize the V2S loss $L$ as defined in (\ref{eq:argmin}), which is evaluated over all target-domain training data. In our implementation of Algorithm \ref{V2S_training} we use stochastic optimization with minibatches.

\begin{algorithm}
    \caption{Voice to Series (V2S) Reprogramming}\label{V2S_training}
    \begin{algorithmic}[1]
    \STATE \textbf{Inputs}: Pre-trained acoustic model $f_{\cS}$, V2S loss $L$  in (\ref{eq:argmin}), target domain training data $\{x_t^{(i)},y_t^{(i)}\}_{i=1}^{n}$,  mask function $M$, multi-label mapping function $h(\cdot)$, maximum number of iterations $T$, initial learning rate $\alpha$
    \STATE \textbf{Output}: Optimal reprogramming parameters $\theta^*$
    \STATE Initialize $\theta$ randomly; set $t= 0$
        \STATE \#\textbf{Generate reprogrammed data input}\\
        $\mathcal{H}(x^{(i)}_{t};\theta) =\text{Pad}(x^{(i)}_{t}) + M\odot \theta$, $\forall~i=\{1,2,\ldots,n\}$
        \STATE \#\textbf{Compute V2S loss $L$ from equation (\ref{eq:argmin})}
        \\
        $L(\theta)=- \frac{1}{n} \sum_{i=1}^n \log P(y_t^{(i)}|f_{\cS}(\mathcal{H}(x^{(i)}_{t});\theta))$
        \STATE \#\textbf{Solve reprogramming parameters}
        \\
        Use ADAM optimizer to solve for $\theta^*$ based on $L(\theta)$
    \end{algorithmic}
    \label{algo:1}
\end{algorithm}

\section{Population Risk via Reprogramming}
\label{sec_risk}

To provide theoretical justification on the effectiveness of V2S, in what follows we establish a formal population risk analysis and prove that
based on V2S, the population risk of the target task is upper bounded by the sum of the source population risk and the Wasserstein-1 distance between the logit representations of the source data and the reprogrammed target data.
Our analysis matches the intuition that a high-accuracy (low population risk) source model with a better source-target data alignment (small Wasserstein-1 distance) should exhibit better reprogramming performance. In Section \ref{subsec_align}, we show that our derived population risk bound can be used to assess the reprogramming performance of V2S for different source models and target tasks. We also note that our theoretical analysis is not limited to V2S reprogramming. It applies to generic classification tasks.

Using the mathematical notation summarized in Table \ref{tab:notation}, the source model is a pre-trained $K$-way neural network classifier $f_{\cS}(\cdot)=\eta(z_{\cS}(\cdot))$ with a softmax layer $\eta(\cdot)$ as the final model output. We omit the notation of the model parameters in our analysis because reprogramming does not change the pre-trained model parameters. The notation $(x,y)$ is used to describe a data sample $x$ and its one-hot coded label $y$. We will use the subscript $s/t$ to denote source/target data when applicable. 
 For the purpose of analysis, given a neural network classifier $f$, we consider the root mean squared error
 (RMSE) denoted by $\|f(x)-y\|_2$.

To put forth our analysis, we make the following assumptions based on the framework of reprogramming:
\begin{enumerate}
    \item The source risk is  $\epsilon_{\cS}$, that is, $\bbE_{\cD_{\cS}} [\ell (x_s,y_s)] = \epsilon_{\cS}$.
    \item The source-target label space has a specified surjective one-to-one label mapping function $h_t$ for every target label $t$, such that $\forall y_t \in \cY_{\cT}$, $y_t=h_t(\cY_{\cS}) \triangleq y_s \in \cY_{\cS}$, and $h_t \neq h_{t'}$ if $t \neq t'$.
    \item Based on reprogramming, the target loss function $\ell_{\cT}$
   with an additive input transformation function $\delta$ can be represented as  $\ell_{\cT}(x_t+\delta,y_t) \overset{(a)}{=} \ell_{\cT}(x_t+\delta,y_s) \overset{(b)}{=} \ell_{\cS}(x_t+\delta,y_s)$, where $(a)$ is induced by label mapping (Assumption 2) and $(b)$ is induced by reprogramming the source loss with target data.
    \item The learned input transformation function for reprogramming is denoted by $\delta^* \triangleq \arg \min_\delta \bbE_{\cD_{\cT}} [\ell_{\cS}(x_t+\delta,y_s)] $, which is the minimizer of the target population risk with the reprogramming loss objective.  
    \item 
    Domain-independent drawing of source and target data: Let $\Phi_{\cS}(\cdot)$ and $\Phi_{\cT}(\cdot)$ denote the probability density function of source data  and target data distributions over $\cX_{\cS}$ and $\cX_{\cT}$, respectively.  The joint probability density function is the product of their marginals, i.e., $\Phi_{\cS,\cT}(x_s,x_t)=\Phi_{\cS}(x_s) \cdot \Phi_{\cT}(x_t)$.
\end{enumerate}

\begin{figure*}[t]
\centering
\includegraphics[width=0.97\textwidth]{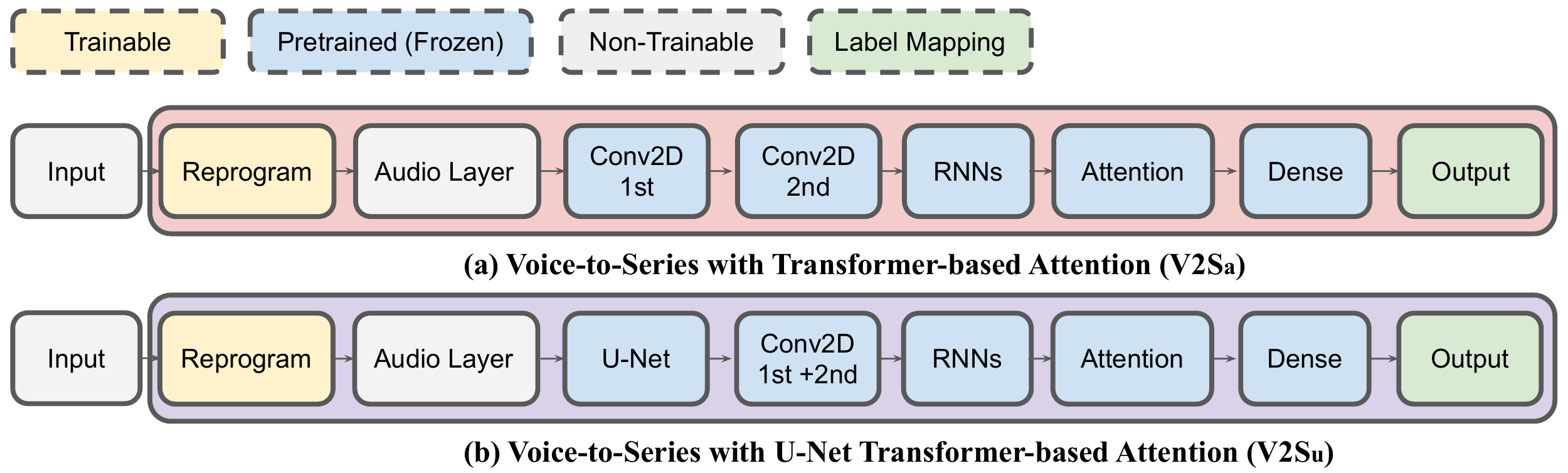}
\caption{V2S architectures: (a) V2S$_a$~\cite{de2018neural} and (b) V2S$_u$ \cite{yang2020decentralizing}. }
\label{fig:tfs:am}
\end{figure*}

For a given neural network classifier, the following lemma associates the expected RMSE of model predictions on two different domains with the Wasserstein-1 distance between their corresponding probability measures on the logit representations, which will play a key role in characterizing the population risk for reprogramming.
Wasserstein distance is a statistical distance between two probability measures $\mu$ and $\mu'$, and it has been widely used for studying optimal transport problems \cite{peyre2018computational}. Specifically, for any $p \geq 1$, the Wasserstein-$p$ distance is defined as
\begin{align*}
    \cW_p(\mu,\mu')= \left( \inf_{\pi \in \Pi (\mu,\mu')} \int \| x-x'\|^p d \pi (x,x') \right)^{1/p},
\end{align*}
where $\Pi (\mu,\mu')$ denotes all joint distributions $\pi$ that have marginals $\mu$ and $\mu'$.

\textbf{Lemma 1:} Given a $K$-way neural network classifier $f(\cdot)=\eta(z(\cdot))$.  Let $\mu_z$ and $\mu_z'$ be the probability measures of the logit representations $\{z(x)\}$ and $\{z(x')\}$ from two data domains $ \cD$ and $ \cD'$, where $x \sim \cD$ and $x' \sim \cD'$. Assume independent draws for $x$ and $x'$, i.e., $\Phi_{\cD,\cD'}(x,x')=\Phi_{\cD}(x) \cdot \Phi_{\cD'}(x')$.
Then 
\begin{align}
    \bbE_{x \sim \cD,~x' \sim \cD'}\|f(x) - f(x') \|_2 \leq 2\sqrt{K} \cdot \cW_1(\mu_z,\mu_z'), \nonumber
\end{align}
 where $\cW_1(\mu_z,\mu_z')$ is the Wasserstein-1 distance between $\mu_z$ and $\mu_z'$.
\\
{\textbf{Proof:} Please see Appendix \ref{proof_lemma}. 

With Lemma 1, we now state the main theorem regarding an upper bound on population risk for reprogramming.

\textbf{Theorem 1:} Let $\delta^*$ denote the learned additive input transformation for reprogramming (Assumption 4).
The population risk for the target task via reprogramming a $K$-way source neural network classifier $f_{\cS}(\cdot)=\eta(z_{\cS}(\cdot))$,  denoted by $\bbE_{\cD_{\cT}}[\ell_{\cT}(x_t+\delta^*,y_t)]$, is upper bounded by
\begin{align*}
&\bbE_{\cD_{\cT}}[\ell_{\cT}(x_t+\delta^*,y_t)] \leq \underbrace{\epsilon_{\cS}}_{\text{source~risk}}\\
&~+
     2\sqrt{K} \cdot \underbrace{\cW_1(\mu(z_{\cS}(x_t+\delta^*)),\mu(z_{\cS}(x_s)))_{x_t \sim \cD_{\cT},~x_s \sim \cD_{\cS}}}_{\text{representation~alignment~loss~via~reprogramming}}
\end{align*}
{\textbf{Proof:} Please see Appendix \ref{proof_thm}.

Theorem 1 shows that the target population risk via reprogramming is upper bounded by the summation of two terms: (i) the source population risk $\epsilon_{\cS}$, and (ii) the representation alignment loss in the logit layer between the source data $z_{\cS}(x_s)$ and the reprogrammed target data $z_{\cS}(x_t+\delta^*)$ based on the same source neural network classifier $f_{\cS}(\cdot) = \eta (z_{\cS}(\cdot))$, measured by their Wasserstein-1 distance. The results suggest that reprogramming can attain better performance (lower risk) when the source model has a lower source loss and a smaller representation alignment loss. 

In the extreme case, if the source and target representations can be fully aligned, the Wasserstein-1 distance will become 0 and thus the target task via reprogramming can perform as well as the source task. On the other hand, if the representation alignment loss is large, then it may dominate the source risk and hinder the performance on the target task. In the next section we will investigate how the representation alignment loss can inform the reprogramming performance for V2S. We would also like to make a final remark that our risk analysis can be extended beyond the additive input transformation setting, by considering a more complex function input transformation function $g(x_t)$  (e.g., an affine transformation). However, in practice, we observe little gain of doing so in V2S and therefore focus on the additive input transformation setting.

\section{Performance Evaluation}
\label{sec_exp}

\subsection{Acoustic Models (AMs) and Source Datasets}
We start by introducing the source models and the source datasets they trained on. The pre-trained source models will be used in our V2S experiments.

\textbf{Limited-vocabulary Voice Commends Dataset:} To create a large-scale ($\sim$100k training samples) pre-trained acoustic model for our experiments, we select the Google Speech Commands V2~\cite{warden2018speech} (denoted as GSCv2) dataset, which contains 105,829 utterances of 35 words from 2,618 recorded speakers with a sampling rate of 16 kHz. 
We also provide some discussion on models trained on other acoustic datasets (e.g., AudioSet~\cite{gemmeke2017audio}) in Appendix~\ref{sub:sect:d}. 
In general, using the same network architecture we find that AMs trained on the voice commands dataset show better V2S performance than other datasets, which could be attributed to its similarity to short-length (one-second or less) time series data.

\textbf{Attention based AMs:}
For training the source model, we use a popular transformer based single-head attention architecture~\cite{de2018neural} for V2S reprogramming, denoted as V2S$_a$ (Figure~\ref{fig:tfs:am} (a)). We also train a similar architecture with  U-Net~\cite{long2015fully}, denoted as V2S$_u$ (Figure~\ref{fig:tfs:am} (b)), which is designed to enhance feature extraction in acoustic  tasks~\cite{yang2020decentralizing}. 
Both pretrained V2S$_a$ and V2S$_u$ models have a comparable number
($\sim$0.2M/0.3M) of model parameters and test accuracy (96.90\%/96.92\%). Their data input dimension is $d\sim$16k and thus the reprogramming function $\theta$ has $\sim$16k trainable parameters.  
The details of the attention based models and comparisons to other popular neural network architectures are given in Appendix \ref{sub:sect:c}.

\begin{table*}[ht!]
\centering
\caption{Performance comparison of test accuracy (\%) on 30 UCR time series classification datasets \cite{dau2019ucr}. Our proposed V2S$_a$ outperforms or ties with the current prediction results (discussed in Section~\ref{sec:sota}) on \textbf{19} out of 30 datasets.}
\label{tab:ucr:1}
\begin{adjustbox}{width=0.98\textwidth}
\begin{tabular}{|l|c|c|c|c|c|c|c|c|}
\hline
Dataset & Type & Input size & Train. Data & Class & SOTA & V2S$_a$ & V2S$_u$ & TF$_a$ \\ \hline \hline
Coffee                               & SPECTRO                & 286                             & 28                               & 2                          & \textbf{100}              & \textbf{100}                 & \textbf{100}               & 53.57                        \\ \hline
DistalPhalanxTW                      & IMAGE                  & 80                              & 400                              & 6                          & \textbf{79.28}            & 79.14               & 75.34                      & 70.21                        \\ \hline
ECG 200                              & ECG                    & 96                              & 100                              & 2                          & \textbf{90.9}                      & 87                 & 87.40               & 81                 \\ \hline
ECG 5000                             & ECG                    & 140                             & 500                              & 5                          & \textbf{94.62}            & 93.96                        & 93.11                      & 58.37                        \\ \hline
Earthquakes                          & SENSOR                 & 512                             & 322                              & 2                          & 76.91                     & \textbf{78.42}               & 76.45                      & 74.82                        \\ \hline
FordA                                & SENSOR                 & 500                             & 2500                             & 2                          & 96.44                     & \textbf{100}                 & \textbf{100}               & \textbf{100}                 \\ \hline
FordB                                & SENSOR                 & 500                             & 3636                             & 2                          & 92.86                     & \textbf{100}                 & \textbf{100}               & \textbf{100}                 \\ \hline
GunPoint                             & MOTION                 & 150                             & 50                               & 2                          & \textbf{100}              & 96.67                        & 93.33                      & 49.33                        \\ \hline
HAM                                  & SPECTROM               & 431                             & 109                              & 2                          & \textbf{83.6}             & 78.1                         & 71.43                      & 51.42                        \\ \hline
HandOutlines                         & IMAGE                  & 2709                            & 1000                             & 2                          & \textbf{93.24}            & \textbf{93.24}               & 91.08                      & 64.05                        \\ \hline
Haptics                              & MOTION                 & 1092                            & 155                              & 5                          & 51.95                     & \textbf{52.27}               & 50.32                      & 21.75                        \\ \hline
Herring                              & IMAGE                  & 512                             & 64                               & 2                          & \textbf{68.75}            & \textbf{68.75}               & 64.06                      & 59.37                        \\ \hline
ItalyPowerDemand                     & SENSOR                 & 24                              & 67                               & 2                          & 97.06                     & \textbf{97.08}               & 96.31                      & 97                           \\ \hline
Lightning2                           & SENSOR                 & 637                             & 60                               & 2                          & 86.89                     & \textbf{100}                 & \textbf{100}               & \textbf{100}                 \\ \hline
MiddlePhalanxOutlineCorrect          & IMAGE                  & 80                              & 600                              & 2                          & 72.23                     & \textbf{83.51}               & 81.79                      & 57.04                        \\ \hline
MiddlePhalanxTW                      & IMAGE                  & 80                              & 399                              & 6                          & 58.69                     & \textbf{65.58}               & 63.64                      & 27.27                        \\ \hline
Plane                                & SENSOR                 & 144                             & 105                              & 7                          & \textbf{100}              & \textbf{100}                 & \textbf{100}               & 9.52                         \\ \hline
ProximalPhalanxOutlineAgeGroup       & IMAGE                  & 80                              & 400                              & 3                          & 88.09                     & \textbf{88.78}               & 87.8                       & 48.78                        \\ \hline
ProximalPhalanxOutlineCorrect        & IMAGE                  & 80                              & 600                              & 2                          & \textbf{92.1}             & 91.07                & 90.03                      & 68.38                        \\ \hline
ProximalPhalanxTW                    & IMAGE                  & 80                              & 400                              & 6                          & 81.86                     & \textbf{84.88}               & 83.41                      & 35.12                        \\ \hline
SmallKitchenAppliances               & DEVICE                 & 720                             & 375                              & 3                          & \textbf{85.33}            & 83.47                        & 74.93                      & 33.33                        \\ \hline
SonyAIBORobotSurface                 & SENSOR                 & 70                              & 20                               & 2                          & \textbf{96.02}            & \textbf{96.02}               & 91.71                      & 34.23                        \\ \hline
Strawberry                           & SPECTRO                & 235                             & 613                              & 2                          & \textbf{98.1}             & 97.57                        & 91.89                      & 64.32                        \\ \hline
SyntheticControl                     & SIMULATED              & 60                              & 300                              & 6                          & \textbf{100}              & 98                           & 99                         & 49.33                        \\ \hline
Trace                                & SENSOR                 & 271                             & 100                              & 4                          & \textbf{100}              & \textbf{100}                 & \textbf{100}               & 18.99                        \\ \hline
TwoLeadECG                           & ECG                    & 82                              & 23                               & 2                          & \textbf{100}              & 96.66                        & 97.81                      & 49.95                        \\ \hline
Wafer                                & SENSOR                 & 152                             & 1000                             & 2                          & 99.98                     & \textbf{100}                 & \textbf{100}               & \textbf{100}                          \\ \hline
WormsTwoClass                        & MOTION                 & 900                             & 181                              & 2                          & 83.12                     & \textbf{98.7}                & 90.91                      & 57.14                        \\ \hline
Worms                                & MOTION                 & 900                             & 181                              & 5                          & 80.17                     & \textbf{83.12}               & 80.34                      & 42.85                        \\ \hline
Wine                                 & SPECTRO                & 234                             & 57                               & 2                          & \textbf{92.61}            & 90.74                        & 90.74                      & 50                           \\ \hline \hline
\textit{Mean accuracy} ($\uparrow$)              & \multicolumn{1}{c|}{-} & \multicolumn{1}{c|}{-}          & \multicolumn{1}{c|}{-}           & \multicolumn{1}{c|}{-}     & 88.02              & \textbf{89.37}         & 87.36                     & 57.57                 \\ \hline
\textit{Median accuracy} ($\uparrow$)            & \multicolumn{1}{c|}{-} & \multicolumn{1}{c|}{-}          & \multicolumn{1}{c|}{-}           & \multicolumn{1}{c|}{-}     & 92.36                     & \textbf{93.60}               & 91.00                      & 55.30                        \\ \hline
\textit{MPCE (mean per class error)} ($\downarrow$) & \multicolumn{1}{c|}{-} & \multicolumn{1}{c|}{-}          & \multicolumn{1}{c|}{-}           & \multicolumn{1}{c|}{-}     & 2.09                      & \textbf{2.04 }                        & 2.13                       & 48.36                        \\ \hline
\end{tabular}
\end{adjustbox}
\end{table*}

\subsection{V2S Implementation and Baseline}
\label{subsec_implement}

\textbf{V2S Implementation:} We use Tensorflow~\cite{abadi2016tensorflow} (v2.2) to implement our V2S framework following Algorithm \ref{V2S_training}. To enable end-to-end V2S training, we use the Kapre toolkit~\cite{choi2017kapre} to incorporate an on-GPU audio preprocessing layer, as shown in Figure~\ref{fig:tfs:am}. 
For the V2S parameters in Algorithm~\ref{V2S_training}, we use $\alpha=0.05$ and a mini-batch size of 32 with  $T=100$ training epochs. We use maximal many-to-one random label mapping, which assigns $\lfloor \frac{|\cY_{\cS}|}{|\cY_{\cT}|} \rfloor$ non-overlapping source labels to every target label, where $|\cY|$ is the size of the label set $\cY$ and $\lfloor z \rfloor$ is the floor function that gives the largest integer not exceeding $z$.
To stabilize the training process, we add weight decay as a regularization term to the V2S loss and set the regularization coefficient to be $0.04$. Our implementation of V2S reprogramming layer is available at \url{https://github.com/huckiyang/Voice2Series-Reprogramming}.

For \emph{model tuning}, we use dropout during training on the reprogramming parameters $\theta$. Moreover, during input reprogramming we also replicate the target signal $x_t$ into $m$ segments and place them starting from the beginning of the reprogrammed input with an identical interval (see Figure \ref{fig:cam} (a) as an example with $m=3$). For each task, we report the best result of V2S among a set of hyperparmeters with dropout rate $\in \{0,0.1,0.2,0.3,0.4\}$ and the number of target signal replication $m \in \{1,2,\ldots,10\}$. We use 10-fold splitting on training data to select the best performed model based on the validation loss and report the accuracy on test data with an average of 10 runs, which follows a similar experimental setting used in~\cite{cabellofast}.

\textbf{Transfer Learning Baseline (TF$_a$):} To demonstrate the effectivness of reprogramming,
we also provide a transfer learning baseline using the same V2S$_a$ pre-trained model. Different from V2S, this baseline (named TF$_a$) does not use input reprogramming but instead allows fine-tuning the pre-trained model parameters using the zero-padded target data. An additional dense layer for task-dependent classification is also included for training.

\subsection{UCR Time Series Classification Benchmark}
\label{sec:sota}
UCR Archive~\cite{dau2019ucr} is a prominent benchmark that contains a large collection of time series classification datasets with default training and testing data splitting. 
 The current state-of-the-art (SOTA) results on test accuracy are obtained from the following methods: (i) deep neural networks including fully convolutional networks (FCN) and deep residual neural networks as reported in~\cite{wang2017time}; (ii) bag-of-features framework~\cite{schafer2015boss, cabellofast}; (iii) ensemble-based framework~\cite{hills2014classification, bagnall2015time,lines2018time}; (iv) time warping framework~\cite{ratanamahatana2005three}.

To ensure each target label is at least assigned with 3 unique source labels, we select 30 time series datasets in UCR Archive with the number of target labels $\leq$ 10 for our V2S experiments. We note that for each dataset, the algorithm that achieves the current result can vary, and therefore the comparison can be unfair to V2S. %

In addition to comparing the standard test accuracy of each dataset as well as the mean and median accuracy over all datasets, we also report the  mean per-class error (MPCE) proposed in~\cite{wang2017time}, which
is a single metric for performance evaluation over multiple datasets.
MPCE is the sum of per-class error (PCE) over $J$ datasets, definded as $\text{MPCE}=\sum_{j \in [J]}\text{PCE}_{j}=\frac{e_{j}}{c_{j}}$, which comprises of the error rate ($e_{j}$) and the number of classes ($c_{j}$) for each dataset.

\textbf{Reprogramming Performance:} Table~\ref{tab:ucr:1} summarizes the performance of each method on 30 datasets.
Notably, our reprogrammed V2S$_a$ model attains either better or equivalent results on \textbf{19} out of 30 time series datasets, suggesting that V2S as a single method is a competitive and promising approach for time series classification.
The transfer learning baseline TF$_a$ has poor performance, which can be attributed to limited training data. V2S$_a$ has higher mean/median accuracy (accuracy increases by 1.84/2.63\%) and lower MPCE (relative error decreases by about 2.87\%) than that of current results, demonstrating the effectiveness of V2S. For most datasets, V2S$_a$ has better performance than V2S$_u$, which can be explained by Theorem 1 through a lower empirical target risk upper bound (see Section \ref{subsec_align}).

\subsection{Representation Alignment Loss} 
\label{subsec_align}

According to Theorem 1, the target risk is upper bounded by the sum of a fixed source risk and a representation alignment loss between the source and reprogrammed target data. The latter is measured by the Wasserstein-1 distance of their logit representations. We use the following experiments to empirically verify the  representation alignment loss during V2S training, and motivate its use for reprogramming performance assessment.
Specifically, for computational efficiency we use the sliced Wasserstein-2 distance (SWD) ~\cite{kolouri2018sliced} to approximate the Wasserstein-1 distance in Theorem 1. SWD uses one-dimensional (1D) random projection (we use 1,000 runs) to compute the sliced Wasserstein-2 distance by invoking 1D-optimal transport (OT), which possesses computational efficiency when compared to higher-dimensional OT problems \cite{peyre2018computational}. Moreover, the Wasserstein-1 distance is upper bounded by the Wasserstein-2 distance \cite{peyre2018computational}, and therefore the SWD will serve as a good approximation of the exact representation alignment loss.

\textbf{Wasserstein Distance during Training:}
Using the DistalPhalanxTW~\cite{davis2013predictive} dataset and V2S$_a$ in Table \ref{tab:ucr:1}, Figure~\ref{fig:2:w:dis} shows the  validation (test) accuracy, validation (test) loss, and SWD during V2S training. One can observe a similar trend between test loss and SWD, suggesting that V2S indeed learns to reprogram the target data representations by gradually  making them closer to the source data distribution, as indicated by Theorem 1.

\begin{figure}[h]
\centering
\includegraphics[width=0.48\textwidth]{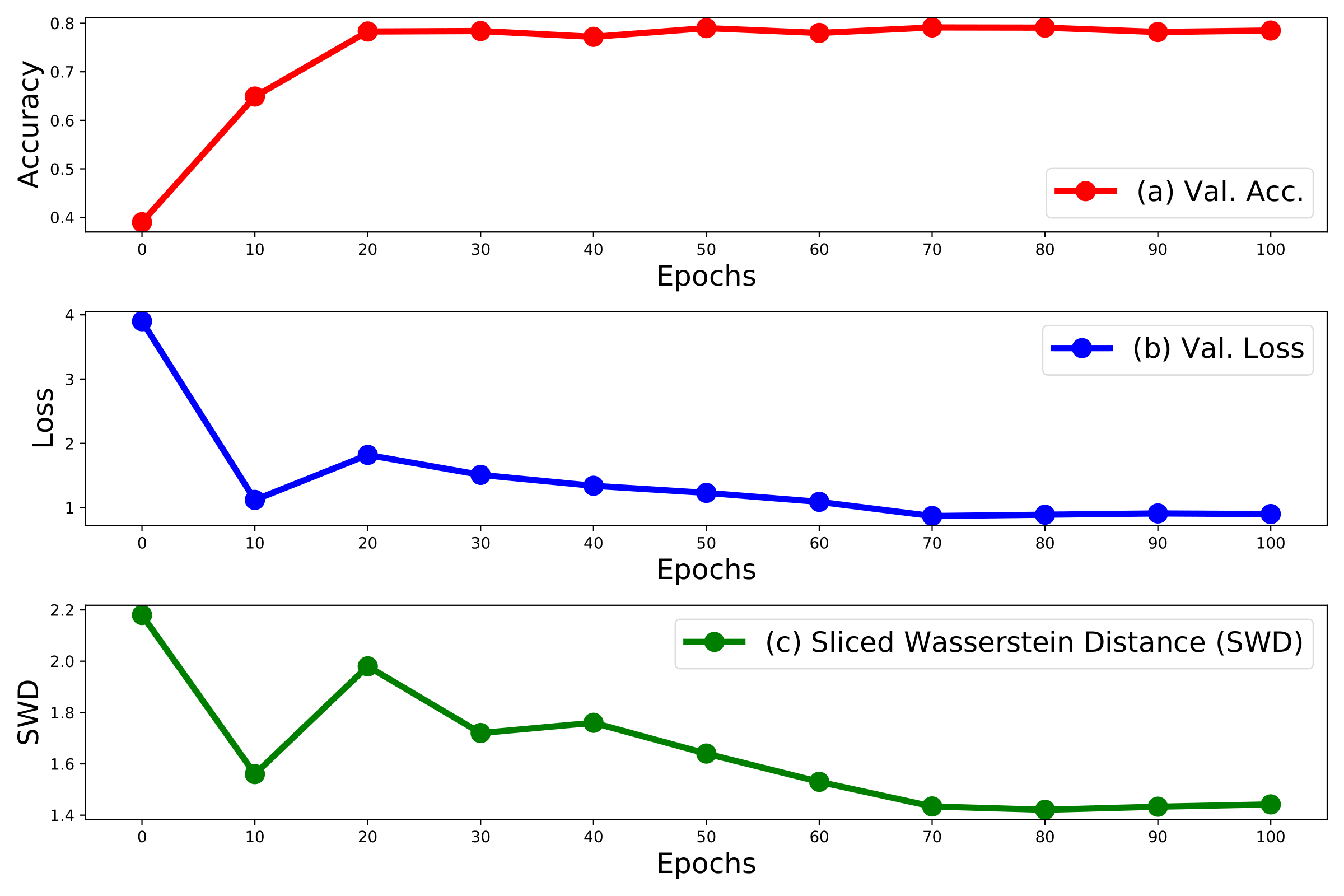}
\vspace{-6mm}
\caption{Training-time reprogramming analysis using V2S$_a$ and DistalPhalanxTW dataset~\cite{davis2013predictive}. All values are averaged over the test set. The rows are (a) validation (test) accuracy, (b) validation loss, and (c) sliced Wasserstein distance  (SWD)~\cite{kolouri2018sliced}.  }
\label{fig:2:w:dis}
\end{figure}

\textbf{Model Selection:} Based on Theorem 1, one can leverage our derived risk bound for V2S model selection.
Comparing V2S$_a$ and V2S$_u$, Table~\ref{tab:w:dis} shows the validation loss of the source task (GSCv2 voice dataset~\cite{warden2018speech}) and the mean/median SWD over all 30 training sets of the target tasks in Table \ref{tab:ucr:1}.
We find that V2S$_a$ indeed has a lower sum of the source loss and SWD than V2S$_u$, which explains its improved performance in Table \ref{tab:ucr:1}.

\begin{table}[t]
\centering
\caption{ Validation loss (Loss$_{\cS}$) of the source task (GSCv2 voice dataset~\cite{warden2018speech}) and  mean/median 
Sliced Wasserstein Distance (SWD) of all training sets in Table \ref{tab:ucr:1}. }
\label{tab:w:dis}
\begin{adjustbox}{width=0.48\textwidth}

\begin{tabular}{|c|c|c|c|c|}
\hline
Model & Loss$_{\cS}$ ($\downarrow$)& Mean SWD ($\downarrow$) & Median SWD ($\downarrow$)\\ \hline
V2S$_a$ & \textbf{0.1709} & \textbf{1.829} & \textbf{1.943} \\ \hline
V2S$_u$ & 0.1734  & 1.873 & 1.977 \\ \hline
\end{tabular}
\end{adjustbox}
\end{table}

\subsection{Additional Analysis on V2S Interpretation}
\label{subsec_visual}

\begin{figure}[ht!]
	\centering
	\begin{subfigure}{0.42\textwidth} %
	    \centering
	    \includegraphics[width=\textwidth]{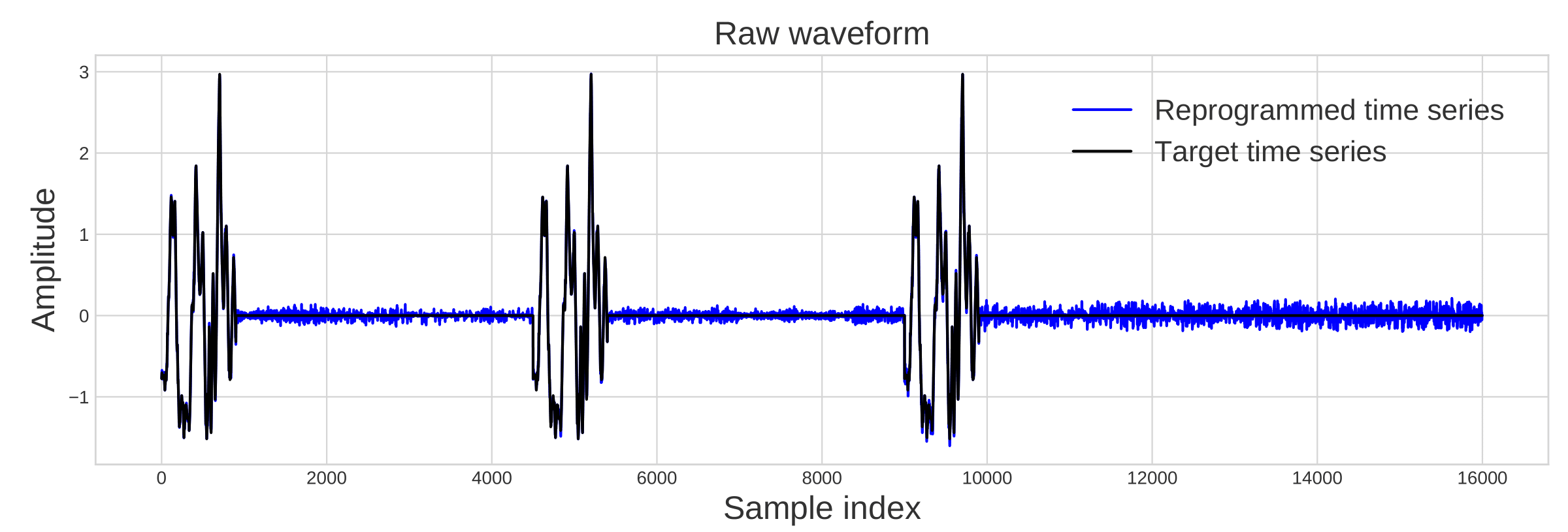}
	    \caption[]%
            {{\small Targeted (blue) and reprogrammed (black) time series}}
	\end{subfigure}
	\quad
	\begin{subfigure}{0.42\textwidth} %
	    \centering
		\includegraphics[width=\textwidth]{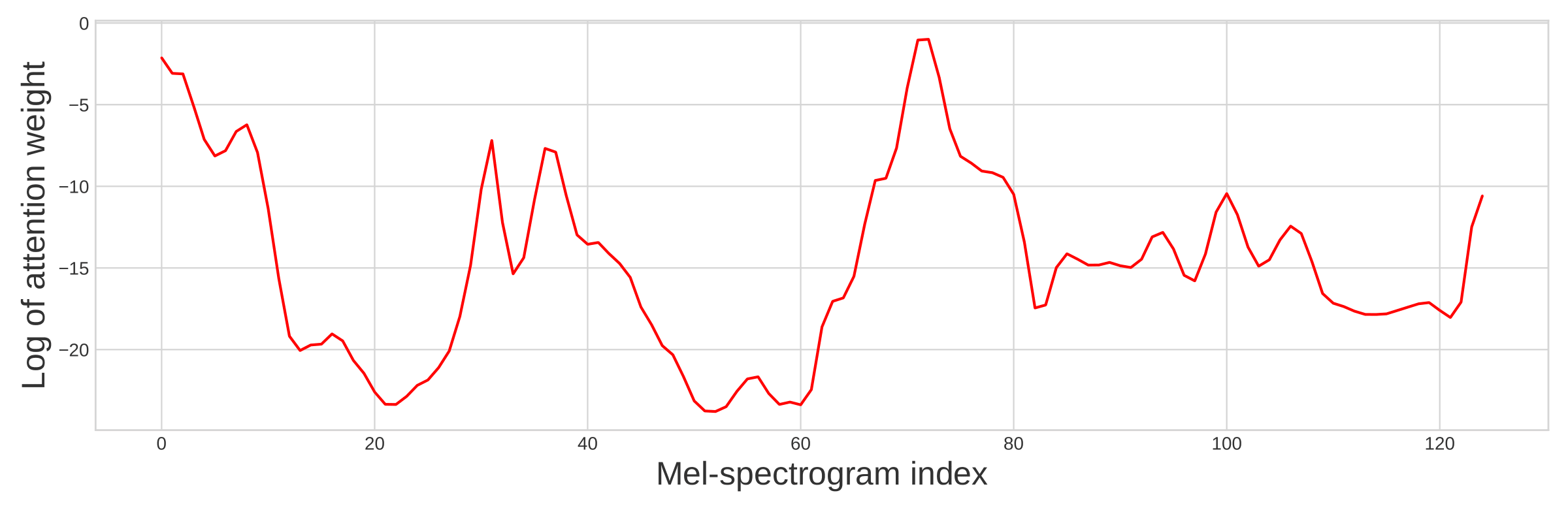}
		\caption[]%
            {{\small Attention weight of reprogrammed input}}
	\end{subfigure}
	\quad
	\begin{subfigure}{0.42\textwidth} %
	    \centering
		\includegraphics[width=\textwidth]{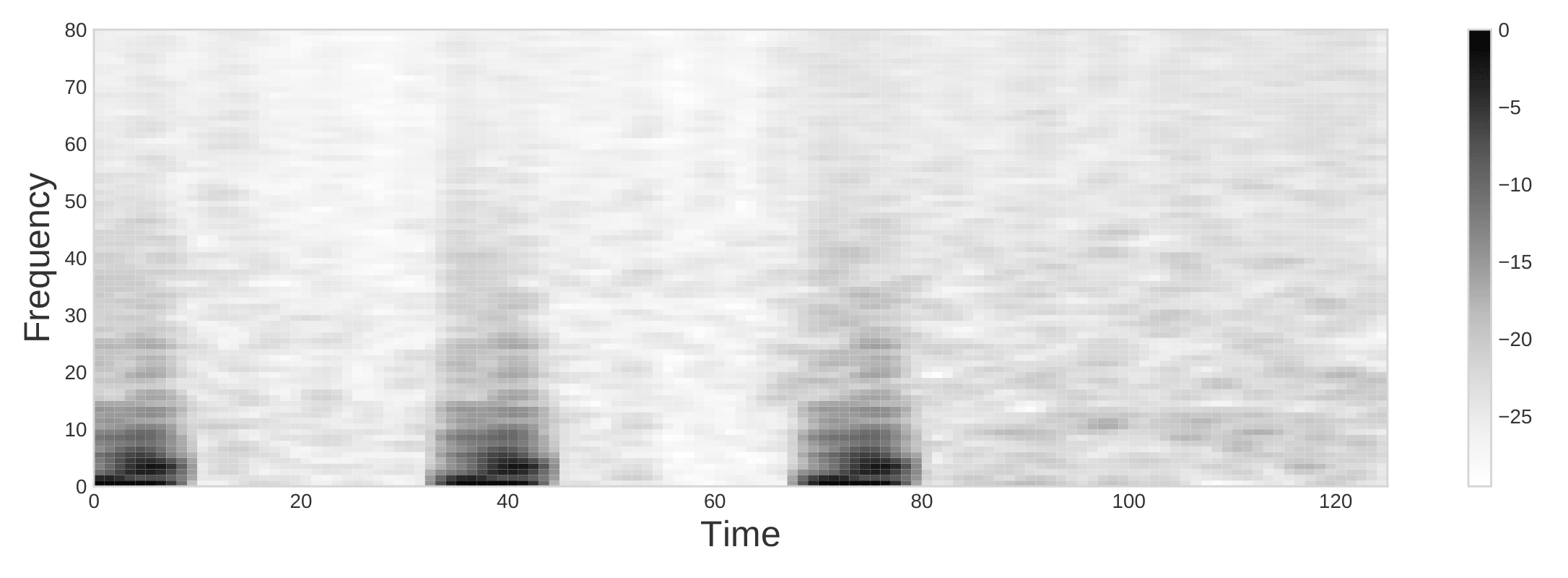}
		\caption[]%
            {{\small Mel-spectrogram of reprogrammed input }}
	\end{subfigure}
	\quad
	\begin{subfigure}{0.42\textwidth} %
	    \centering
		\includegraphics[width=\textwidth]{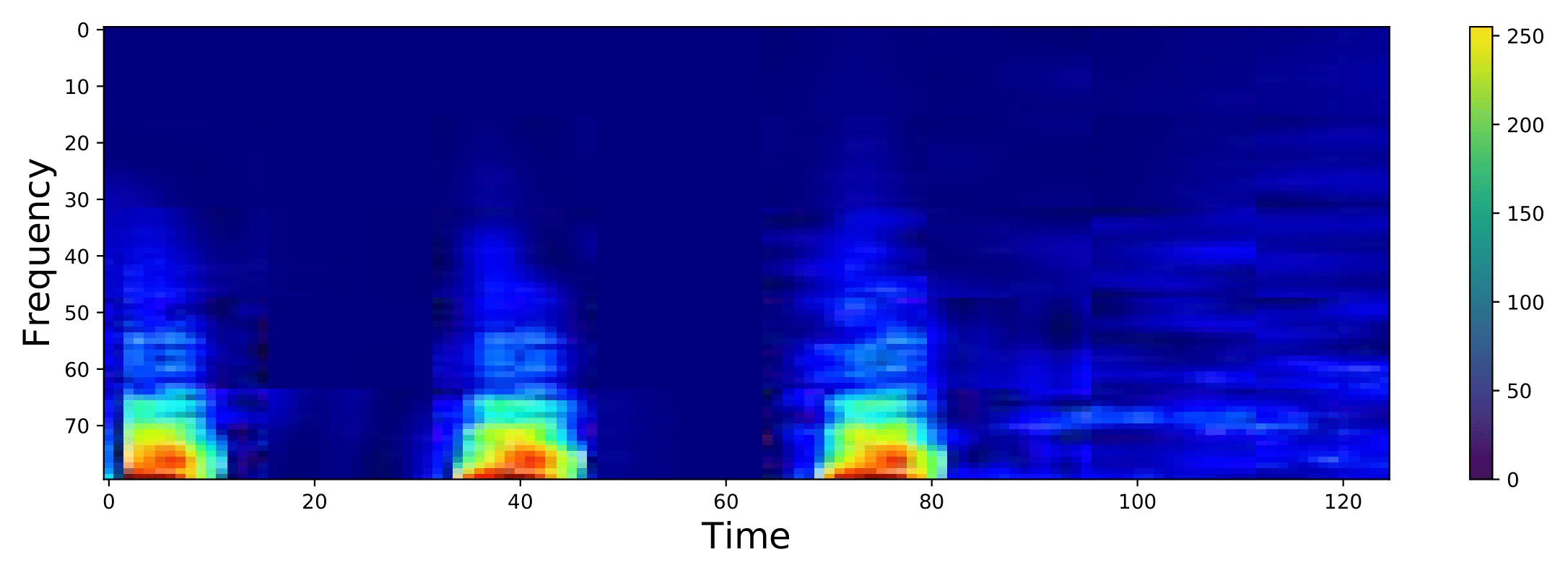}
		\caption[]%
            {{\small Class activation mapping of (c) from 1$_\text{st}$ conv-layer }}
	\end{subfigure}
	\quad
	\begin{subfigure}{0.42\textwidth} %
	    \centering
		\includegraphics[width=\textwidth]{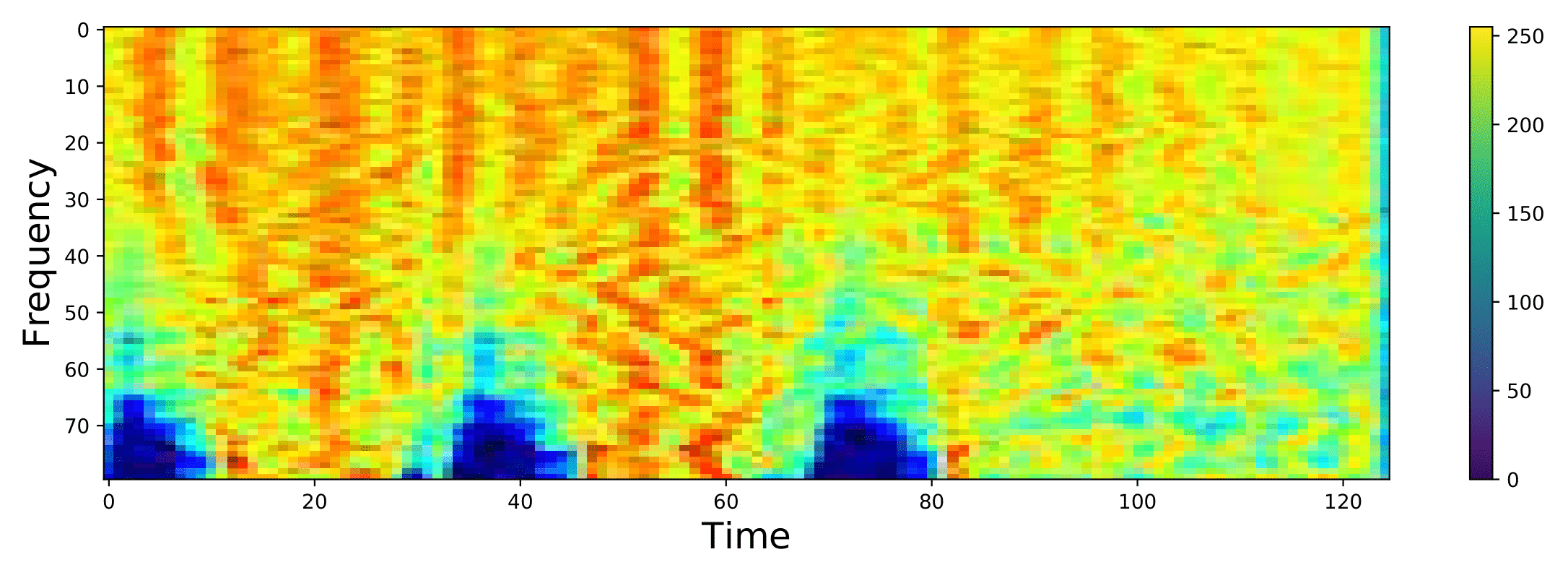}
		\caption[]%
            {{\small Class activation mapping of (c) from 2$_\text{nd}$ conv-layer }}
	\end{subfigure}
\vspace{-0.2cm}
   \caption{ Visualization of a data sample in the Worms dataset~\cite{bagnall2015time} using V2S$_a$. The rows are (a) the original target time series and its reprogrammed pattern as illustrated in Figure~\ref{fig:1:sys}, (b) the associated attention-head predicted by V2S$_a$, (c) Mel-spectrogram of the reprogrammed input, and (d)/(e) its neural saliency maps via class activation mapping~\cite{zhou2016learning} from the first/second convolution layer.} 
\label{fig:cam}
\end{figure}

\begin{figure*}[t]
\centering
\includegraphics[width=0.97\textwidth]{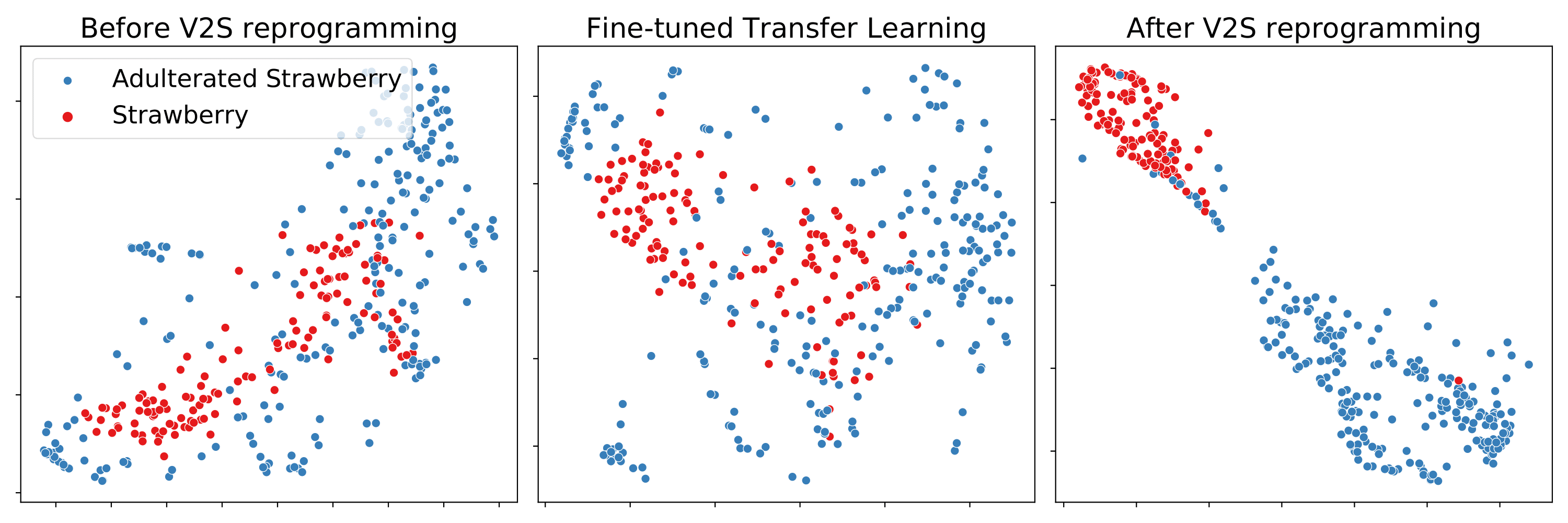}
\caption{tSNE plots of the logit representations using the Strawberry training set~\cite{holland1998use} and V2S$_a$, for the cases of before and after V2S reprogramming, and fine-tuned transfer learning (TF$_a$). }
\label{fig:3:tsne}
\end{figure*}

To gain further insights on V2S, we study its acoustic saliency map and embedding visualization. 

\textbf{Attention and Class Activation Mapping:}
To interpret the prediction made by V2S, we provide neural saliency analysis over the spectrogram of the reprogrammed features by class activation mapping (CAM)~\cite{zhou2016learning} using the Worms dataset~\cite{bagnall2015time} and V2S$_a$. Activation and attention mapping methods~\cite{wu2019enhancing} have been used in auditory analysis~\cite{fritz2007auditory} and investigated on its relationship between audio signal and brain cortex activation by neural physiology studies~\cite{kaya2017modelling, veale2017visual}. Interestingly, as shown in Figure~\ref{fig:cam}, the corresponding attention head (b) of pre-trained AM (non-trainable during V2S process) could still recognize the original temporal patterns from the reprogrammed input signal in (a). We also show the Mel-spectrogram of reprogrammed input in (c), indicating the activated spatial-temporal acoustic features corresponding to weighted output prediction. Furthermore, 
in reference to the V2S architecture introduced in Figure~\ref{fig:tfs:am}, we select the first and the second convolution layer for CAM visualization in (d) and (e). From the analysis, we observe different functions of these two layers. The first 
convolution layer tends to focus on the
target signal segments themselves as well as their
low-frequency acoustic features of the reprogrammed input's Mel-spectrogram in (c), whereas the second convolution layer tends to put more focus on the high-frequency components in the reprogrammed input.

\textbf{Embedding Visualization:} 
We use t-distributed stochastic neighbor embedding (tSNE)~\cite{van2008visualizing} to visualize the logit representations of the Strawberry training set~\cite{holland1998use} for the cases of before and after reprogramming, and the transfer learning baseline (TF$_a$).
As shown in Figure~\ref{fig:3:tsne}, after reprogramming tSNE results show a clear separation between the embeddings from different target classes,  suggesting that V2S
 indeed learns meaningful and discriminative data representations to reprogram the pre-trained acoustic model for time series classification. On the other hand, the embedding visualization of transfer learning shows low class-wise separability.

\subsection{Additional Discussion}
In what follows, we provide additional discussion and insights for V2S reprogramming.

\textbf{Many-to-one Label Mapping:} 
The many-to-one mapping can be viewed as ensemble averaging of single-class outputs, a practical technique to improve classification for V2S reprogramming. Our current results show that many-to-one label mapping performs better results in control experiments.

\textbf{Significance Testing:}
In Section \ref{sec_exp}, our results suggest that V2S as a \textit{single} method can achieve or exceed competitive performance on \textbf{19} out of 30 datasets obtained from \textit{different} methods.
We further run significant testing on two levels using their accuracies based on the 30 datasets from Table~\ref{tab:ucr:1} --- (i) V2S$_a$ v.s. SOTA numbers: p-value=0.0017; (ii) V2S v.s. FCN~\cite{wang2017time}: p-value=0.0011, which indicate our results are significant.

\textbf{Effect of Source Dataset}:
The effect of source dataset for V2S is captured by the source risk $\epsilon_{S}$ in Theorem 1, along with the representation alignment loss via SWD. We first evaluate different acoustic datasets to calculate their source test errors,  where GSCv2 attains the smallest test error. The source test errors on $\{$GSCv2, TAU, AudioSet, ESC$\}$ are $\{$0.1709, 0.1822, 0.1839, 0.1765$\}$, and their SWD are reported in Appendix Table~\ref{tab:sub:pre:more}.
Our theorem informs the performance of V2S on different source datasets. If other datasets can have smaller $\epsilon_{S}$ and SWD than GSCv2, we expect it to have better V2S performance.

\textbf{Future Works:}
Our future works include a wider range of performance evaluations on different acoustic and speech models (e.g., those associated with lexical information) for V2S reprogramming, model reprogramming for low-resource speech processing, and extension to multivariate time series tasks. The proposed theory would also provide insights on analyzing the success of adversarial reprogramming in vision and language processing domains.

\section{Conclusion}
In this work, we proposed V2S, a novel approach to reprogram a pre-trained acoustic model for time series classification. We also developed a theoretical risk analysis to characterize the reprogramming performance. 
Experimental results on UCR benchmark showed superior performance of V2S, by achieving new (or equal) state-of-the-art accuracy on \textbf{19} out of 30 datasets. We also provided in-depth studies on the success of V2S through representation alignment, acoustic saliency map, and embedding visualization. The V2S could further incorporate with different advanced data augmentation techniques for future studies. 

\section*{Acknowledgements}
The authors would like to thank the comments and discussion from anonymous reviewers during the double-blind review process. 

\clearpage

\bibliography{example_paper,adversarial_learning}

\begin{thebibliography}{60}
\providecommand{\natexlab}[1]{#1}
\providecommand{\url}[1]{\texttt{#1}}
\expandafter\ifx\csname urlstyle\endcsname\relax
  \providecommand{\doi}[1]{doi: #1}\else
  \providecommand{\doi}{doi: \begingroup \urlstyle{rm}\Url}\fi

\bibitem[Abadi et~al.(2016)Abadi, Barham, Chen, Chen, Davis, Dean, Devin,
  Ghemawat, Irving, Isard, et~al.]{abadi2016tensorflow}
Abadi, M., Barham, P., Chen, J., Chen, Z., Davis, A., Dean, J., Devin, M.,
  Ghemawat, S., Irving, G., Isard, M., et~al.
\newblock Tensorflow: A system for large-scale machine learning.
\newblock In \emph{12th $\{$USENIX$\}$ symposium on operating systems design
  and implementation ($\{$OSDI$\}$ 16)}, pp.\  265--283, 2016.

\bibitem[Bagnall et~al.(2015)Bagnall, Lines, Hills, and
  Bostrom]{bagnall2015time}
Bagnall, A., Lines, J., Hills, J., and Bostrom, A.
\newblock Time-series classification with cote: the collective of
  transformation-based ensembles.
\newblock \emph{IEEE Transactions on Knowledge and Data Engineering},
  27\penalty0 (9):\penalty0 2522--2535, 2015.

\bibitem[Cabello et~al.(2020)Cabello, Naghizade, Qi, and Kulik]{cabellofast}
Cabello, N., Naghizade, E., Qi, J., and Kulik, L.
\newblock Fast and accurate time series classification through supervised
  interval search.
\newblock \emph{ICDM 2020}, 2020.

\bibitem[Chen et~al.(2017)Chen, Zhang, Sharma, Yi, and Hsieh]{chen2017zoo}
Chen, P.-Y., Zhang, H., Sharma, Y., Yi, J., and Hsieh, C.-J.
\newblock {ZOO}: Zeroth order optimization based black-box attacks to deep
  neural networks without training substitute models.
\newblock In \emph{ACM Workshop on Artificial Intelligence and Security}, pp.\
  15--26, 2017.

\bibitem[Choi et~al.(2017)Choi, Joo, and Kim]{choi2017kapre}
Choi, K., Joo, D., and Kim, J.
\newblock Kapre: On-gpu audio preprocessing layers for a quick implementation
  of deep neural network models with keras.
\newblock In \emph{Machine Learning for Music Discovery Workshop at 34th
  International Conference on Machine Learning}. ICML, 2017.

\bibitem[Cramer et~al.(2019)Cramer, Wu, Salamon, and Bello]{cramer2019look}
Cramer, J., Wu, H.-H., Salamon, J., and Bello, J.~P.
\newblock Look, listen, and learn more: Design choices for deep audio
  embeddings.
\newblock In \emph{ICASSP 2019-2019 IEEE International Conference on Acoustics,
  Speech and Signal Processing (ICASSP)}, pp.\  3852--3856. IEEE, 2019.

\bibitem[Dau et~al.(2019)Dau, Bagnall, Kamgar, Yeh, Zhu, Gharghabi,
  Ratanamahatana, and Keogh]{dau2019ucr}
Dau, H.~A., Bagnall, A., Kamgar, K., Yeh, C.-C.~M., Zhu, Y., Gharghabi, S.,
  Ratanamahatana, C.~A., and Keogh, E.
\newblock The ucr time series archive.
\newblock \emph{IEEE/CAA Journal of Automatica Sinica}, 6\penalty0
  (6):\penalty0 1293--1305, 2019.

\bibitem[Davis(2013)]{davis2013predictive}
Davis, L.~M.
\newblock \emph{Predictive modelling of bone ageing}.
\newblock PhD thesis, University of East Anglia, 2013.

\bibitem[de~Andrade et~al.(2018)de~Andrade, Leo, Viana, and
  Bernkopf]{de2018neural}
de~Andrade, D.~C., Leo, S., Viana, M. L. D.~S., and Bernkopf, C.
\newblock A neural attention model for speech command recognition.
\newblock \emph{arXiv preprint arXiv:1808.08929}, 2018.

\bibitem[Elsayed et~al.(2019)Elsayed, Goodfellow, and
  Sohl-Dickstein]{elsayed2018adversarial}
Elsayed, G.~F., Goodfellow, I., and Sohl-Dickstein, J.
\newblock Adversarial reprogramming of neural networks.
\newblock In \emph{International Conference on Learning Representations}, 2019.

\bibitem[Fawaz et~al.(2018)Fawaz, Forestier, Weber, Idoumghar, and
  Muller]{fawaz2018transfer}
Fawaz, H.~I., Forestier, G., Weber, J., Idoumghar, L., and Muller, P.-A.
\newblock Transfer learning for time series classification.
\newblock In \emph{2018 IEEE international conference on big data (Big Data)},
  pp.\  1367--1376. IEEE, 2018.

\bibitem[Fawaz et~al.(2019)Fawaz, Forestier, Weber, Idoumghar, and
  Muller]{fawaz2019deep}
Fawaz, H.~I., Forestier, G., Weber, J., Idoumghar, L., and Muller, P.-A.
\newblock Deep learning for time series classification: a review.
\newblock \emph{Data Mining and Knowledge Discovery}, 33\penalty0 (4):\penalty0
  917--963, 2019.

\bibitem[Fritz et~al.(2007)Fritz, Elhilali, David, and
  Shamma]{fritz2007auditory}
Fritz, J.~B., Elhilali, M., David, S.~V., and Shamma, S.~A.
\newblock Auditory attention—focusing the searchlight on sound.
\newblock \emph{Current opinion in neurobiology}, 17\penalty0 (4):\penalty0
  437--455, 2007.

\bibitem[Gao \& Pavel(2017)Gao and Pavel]{gao2017properties}
Gao, B. and Pavel, L.
\newblock On the properties of the softmax function with application in game
  theory and reinforcement learning.
\newblock \emph{arXiv preprint arXiv:1704.00805}, 2017.

\bibitem[Gemmeke et~al.(2017)Gemmeke, Ellis, Freedman, Jansen, Lawrence, Moore,
  Plakal, and Ritter]{gemmeke2017audio}
Gemmeke, J.~F., Ellis, D.~P., Freedman, D., Jansen, A., Lawrence, W., Moore,
  R.~C., Plakal, M., and Ritter, M.
\newblock Audio set: An ontology and human-labeled dataset for audio events.
\newblock In \emph{2017 IEEE International Conference on Acoustics, Speech and
  Signal Processing (ICASSP)}, pp.\  776--780. IEEE, 2017.

\bibitem[Geurts(2001)]{geurts2001pattern}
Geurts, P.
\newblock Pattern extraction for time series classification.
\newblock In \emph{European conference on principles of data mining and
  knowledge discovery}, pp.\  115--127. Springer, 2001.

\bibitem[Hambardzumyan et~al.(2020)Hambardzumyan, Khachatrian, and
  May]{hambardzumyan2020warp}
Hambardzumyan, K., Khachatrian, H., and May, J.
\newblock Warp: Word-level adversarial reprogramming.
\newblock \emph{arXiv preprint arXiv:2101.00121}, 2020.

\bibitem[He et~al.(2016)He, Zhang, Ren, and Sun]{he2016deep}
He, K., Zhang, X., Ren, S., and Sun, J.
\newblock Deep residual learning for image recognition.
\newblock In \emph{Proceedings of the IEEE conference on computer vision and
  pattern recognition}, pp.\  770--778, 2016.

\bibitem[Heittola et~al.(2020)Heittola, Mesaros, and
  Virtanen]{heittola2020acoustic}
Heittola, T., Mesaros, A., and Virtanen, T.
\newblock Acoustic scene classification in dcase 2020 challenge: generalization
  across devices and low complexity solutions.
\newblock \emph{arXiv preprint arXiv:2005.14623}, 2020.

\bibitem[Hershey et~al.(2017)Hershey, Chaudhuri, Ellis, Gemmeke, Jansen, Moore,
  Plakal, Platt, Saurous, Seybold, et~al.]{hershey2017cnn}
Hershey, S., Chaudhuri, S., Ellis, D.~P., Gemmeke, J.~F., Jansen, A., Moore,
  R.~C., Plakal, M., Platt, D., Saurous, R.~A., Seybold, B., et~al.
\newblock Cnn architectures for large-scale audio classification.
\newblock In \emph{2017 ieee international conference on acoustics, speech and
  signal processing (icassp)}, pp.\  131--135. IEEE, 2017.

\bibitem[Hills et~al.(2014)Hills, Lines, Baranauskas, Mapp, and
  Bagnall]{hills2014classification}
Hills, J., Lines, J., Baranauskas, E., Mapp, J., and Bagnall, A.
\newblock Classification of time series by shapelet transformation.
\newblock \emph{Data mining and knowledge discovery}, 28\penalty0 (4):\penalty0
  851--881, 2014.

\bibitem[Holland et~al.(1998)Holland, Kemsley, and Wilson]{holland1998use}
Holland, J., Kemsley, E., and Wilson, R.
\newblock Use of fourier transform infrared spectroscopy and partial least
  squares regression for the detection of adulteration of strawberry purees.
\newblock \emph{Journal of the Science of Food and Agriculture}, 76\penalty0
  (2):\penalty0 263--269, 1998.

\bibitem[Hong et~al.(2019)Hong, Xiao, Ma, Li, and Sun]{hong2019mina}
Hong, S., Xiao, C., Ma, T., Li, H., and Sun, J.
\newblock Mina: Multilevel knowledge-guided attention for modeling
  electrocardiography signals.
\newblock \emph{arXiv preprint arXiv:1905.11333}, 2019.

\bibitem[Hu et~al.(2020)Hu, Yang, Xia, Bai, Tang, Wang, Niu, Chai, Li, Zhu,
  et~al.]{hu2020device}
Hu, H., Yang, C.-H.~H., Xia, X., Bai, X., Tang, X., Wang, Y., Niu, S., Chai,
  L., Li, J., Zhu, H., et~al.
\newblock Device-robust acoustic scene classification based on two-stage
  categorization and data augmentation.
\newblock \emph{arXiv preprint arXiv:2007.08389}, 2020.

\bibitem[Hu et~al.(2021)Hu, Yang, Xia, Bai, Tang, Wang, Niu, Chai, Li, Zhu,
  et~al.]{hu2021two}
Hu, H., Yang, C.-H.~H., Xia, X., Bai, X., Tang, X., Wang, Y., Niu, S., Chai,
  L., Li, J., Zhu, H., et~al.
\newblock A two-stage approach to device-robust acoustic scene classification.
\newblock In \emph{ICASSP 2021-2021 IEEE International Conference on Acoustics,
  Speech and Signal Processing (ICASSP)}, pp.\  845--849. IEEE, 2021.

\bibitem[Kampouraki et~al.(2008)Kampouraki, Manis, and
  Nikou]{kampouraki2008heartbeat}
Kampouraki, A., Manis, G., and Nikou, C.
\newblock Heartbeat time series classification with support vector machines.
\newblock \emph{IEEE Transactions on Information Technology in Biomedicine},
  13\penalty0 (4):\penalty0 512--518, 2008.

\bibitem[Kantorovich \& Rubinstein(1958)Kantorovich and
  Rubinstein]{Kantorovich_Rubinstein_thm}
Kantorovich, L. and Rubinstein, G.
\newblock On a space of completely additive functions.
\newblock In \emph{Vestnik Leningradskogo Universiteta}, volume 13 (7), pp.\
  52--59, 1958.

\bibitem[Kashiparekh et~al.(2019)Kashiparekh, Narwariya, Malhotra, Vig, and
  Shroff]{kashiparekh2019convtimenet}
Kashiparekh, K., Narwariya, J., Malhotra, P., Vig, L., and Shroff, G.
\newblock Convtimenet: A pre-trained deep convolutional neural network for time
  series classification.
\newblock In \emph{2019 International Joint Conference on Neural Networks
  (IJCNN)}, pp.\  1--8. IEEE, 2019.

\bibitem[Kaya \& Elhilali(2017)Kaya and Elhilali]{kaya2017modelling}
Kaya, E.~M. and Elhilali, M.
\newblock Modelling auditory attention.
\newblock \emph{Philosophical Transactions of the Royal Society B: Biological
  Sciences}, 372\penalty0 (1714):\penalty0 20160101, 2017.

\bibitem[Kingma \& Ba(2015)Kingma and Ba]{kingma2014adam}
Kingma, D. and Ba, J.
\newblock Adam: A method for stochastic optimization.
\newblock \emph{International Conference on Learning Representations}, 2015.

\bibitem[Kloberdanz(2020)]{kloberdanz2020reprogramming}
Kloberdanz, E.
\newblock Reprogramming of neural networks: A new and improved machine learning
  technique.
\newblock \emph{ISU Master Thesis}, 2020.

\bibitem[Kolouri et~al.(2018)Kolouri, Rohde, and Hoffmann]{kolouri2018sliced}
Kolouri, S., Rohde, G.~K., and Hoffmann, H.
\newblock Sliced wasserstein distance for learning gaussian mixture models.
\newblock In \emph{Proceedings of the IEEE Conference on Computer Vision and
  Pattern Recognition}, pp.\  3427--3436, 2018.

\bibitem[Langkvist et~al.(2014)Langkvist, Karlsson, and
  Loutfi]{langkvist2014review}
Langkvist, M., Karlsson, L., and Loutfi, A.
\newblock A review of unsupervised feature learning and deep learning for
  time-series modeling.
\newblock \emph{Pattern Recognition Letters}, 42:\penalty0 11--24, 2014.

\bibitem[Lea et~al.(2016)Lea, Vidal, Reiter, and Hager]{lea2016temporal}
Lea, C., Vidal, R., Reiter, A., and Hager, G.~D.
\newblock Temporal convolutional networks: A unified approach to action
  segmentation.
\newblock In \emph{European Conference on Computer Vision}, pp.\  47--54.
  Springer, 2016.

\bibitem[Lines et~al.(2012)Lines, Davis, Hills, and Bagnall]{lines2012shapelet}
Lines, J., Davis, L.~M., Hills, J., and Bagnall, A.
\newblock A shapelet transform for time series classification.
\newblock In \emph{Proceedings of the 18th ACM SIGKDD international conference
  on Knowledge discovery and data mining}, pp.\  289--297, 2012.

\bibitem[Lines et~al.(2018)Lines, Taylor, and Bagnall]{lines2018time}
Lines, J., Taylor, S., and Bagnall, A.
\newblock Time series classification with hive-cote: The hierarchical vote
  collective of transformation-based ensembles.
\newblock \emph{ACM Transactions on Knowledge Discovery from Data}, 12\penalty0
  (5), 2018.

\bibitem[Liu et~al.(2020)Liu, Chen, Kailkhura, Zhang, Hero, and
  Varshney]{liu2020primer}
Liu, S., Chen, P.-Y., Kailkhura, B., Zhang, G., Hero, A., and Varshney, P.~K.
\newblock A primer on zeroth-order optimization in signal processing and
  machine learning.
\newblock \emph{IEEE Signal Processing Magazine}, 2020.

\bibitem[Long et~al.(2015)Long, Shelhamer, and Darrell]{long2015fully}
Long, J., Shelhamer, E., and Darrell, T.
\newblock Fully convolutional networks for semantic segmentation.
\newblock In \emph{Proceedings of the IEEE conference on computer vision and
  pattern recognition}, pp.\  3431--3440, 2015.

\bibitem[Neekhara et~al.(2019)Neekhara, Hussain, Dubnov, and
  Koushanfar]{neekhara2019adversarial}
Neekhara, P., Hussain, S., Dubnov, S., and Koushanfar, F.
\newblock Adversarial reprogramming of text classification neural networks.
\newblock In \emph{Proceedings of the 2019 Conference on Empirical Methods in
  Natural Language Processing and the 9th International Joint Conference on
  Natural Language Processing (EMNLP-IJCNLP)}, pp.\  5219--5228, 2019.

\bibitem[Peyr{\'e} \& Cuturi(2018)Peyr{\'e} and Cuturi]{peyre2018computational}
Peyr{\'e}, G. and Cuturi, M.
\newblock Computational optimal transport. arxiv e-prints.
\newblock \emph{arXiv preprint arXiv:1803.00567}, 2018.

\bibitem[Piczak(2025)]{piczak2015dataset}
Piczak, K.~J.
\newblock {ESC}: {Dataset} for {Environmental Sound Classification}.
\newblock In \emph{Proceedings of the 23rd {Annual ACM Conference} on
  {Multimedia}}, pp.\  1015--1018. {ACM Press}, 2025.
\newblock ISBN 978-1-4503-3459-4.
\newblock \doi{10.1145/2733373.2806390}.
\newblock URL \url{http://dl.acm.org/citation.cfm?doid=2733373.2806390}.

\bibitem[Ratanamahatana \& Keogh(2005)Ratanamahatana and
  Keogh]{ratanamahatana2005three}
Ratanamahatana, C.~A. and Keogh, E.
\newblock Three myths about dynamic time warping data mining.
\newblock In \emph{Proceedings of the 2005 SIAM international conference on
  data mining}, pp.\  506--510. SIAM, 2005.

\bibitem[Sandler et~al.(2018)Sandler, Howard, Zhu, Zhmoginov, and
  Chen]{sandler2018mobilenetv2}
Sandler, M., Howard, A., Zhu, M., Zhmoginov, A., and Chen, L.-C.
\newblock Mobilenetv2: Inverted residuals and linear bottlenecks.
\newblock In \emph{Proceedings of the IEEE conference on computer vision and
  pattern recognition}, pp.\  4510--4520, 2018.

\bibitem[Saon et~al.(2017)Saon, Kurata, Sercu, Audhkhasi, Thomas, Dimitriadis,
  Cui, Ramabhadran, Picheny, Lim, et~al.]{saon2017english}
Saon, G., Kurata, G., Sercu, T., Audhkhasi, K., Thomas, S., Dimitriadis, D.,
  Cui, X., Ramabhadran, B., Picheny, M., Lim, L.-L., et~al.
\newblock English conversational telephone speech recognition by humans and
  machines.
\newblock \emph{Proc. Interspeech 2017}, pp.\  132--136, 2017.

\bibitem[Sch{\"a}fer(2015)]{schafer2015boss}
Sch{\"a}fer, P.
\newblock The boss is concerned with time series classification in the presence
  of noise.
\newblock \emph{Data Mining and Knowledge Discovery}, 29\penalty0 (6):\penalty0
  1505--1530, 2015.

\bibitem[Simonyan \& Zisserman(2015)Simonyan and Zisserman]{simonyan2014very}
Simonyan, K. and Zisserman, A.
\newblock Very deep convolutional networks for large-scale image recognition.
\newblock \emph{International Conference on Learning Representations}, 2015.

\bibitem[Tsai et~al.(2020)Tsai, Chen, and Ho]{tsai2020transfer}
Tsai, Y.-Y., Chen, P.-Y., and Ho, T.-Y.
\newblock Transfer learning without knowing: Reprogramming black-box machine
  learning models with scarce data and limited resources.
\newblock In \emph{International Conference on Machine Learning}, pp.\
  9614--9624, 2020.

\bibitem[Van~der Maaten \& Hinton(2008)Van~der Maaten and
  Hinton]{van2008visualizing}
Van~der Maaten, L. and Hinton, G.
\newblock Visualizing data using t-sne.
\newblock \emph{Journal of machine learning research}, 9\penalty0 (11), 2008.

\bibitem[Veale et~al.(2017)Veale, Hafed, and Yoshida]{veale2017visual}
Veale, R., Hafed, Z.~M., and Yoshida, M.
\newblock How is visual salience computed in the brain? insights from
  behaviour, neurobiology and modelling.
\newblock \emph{Philosophical Transactions of the Royal Society B: Biological
  Sciences}, 372\penalty0 (1714):\penalty0 20160113, 2017.

\bibitem[Vinod et~al.(2020)Vinod, Chen, and Das]{vinod2020reprogramming}
Vinod, R., Chen, P.-Y., and Das, P.
\newblock Reprogramming language models for molecular representation learning.
\newblock \emph{arXiv preprint arXiv:2012.03460}, 2020.

\bibitem[Wang et~al.(2017)Wang, Yan, and Oates]{wang2017time}
Wang, Z., Yan, W., and Oates, T.
\newblock Time series classification from scratch with deep neural networks: A
  strong baseline.
\newblock In \emph{2017 International joint conference on neural networks
  (IJCNN)}, pp.\  1578--1585. IEEE, 2017.

\bibitem[Warden(2018)]{warden2018speech}
Warden, P.
\newblock Speech commands: A dataset for limited-vocabulary speech recognition.
\newblock \emph{arXiv preprint arXiv:1804.03209}, 2018.

\bibitem[Wu \& Lee(2019)Wu and Lee]{wu2019enhancing}
Wu, Y. and Lee, T.
\newblock Enhancing sound texture in cnn-based acoustic scene classification.
\newblock In \emph{ICASSP 2019-2019 IEEE International Conference on Acoustics,
  Speech and Signal Processing (ICASSP)}, pp.\  815--819. IEEE, 2019.

\bibitem[Xu et~al.(2014)Xu, Du, Dai, and Lee]{xu2014regression}
Xu, Y., Du, J., Dai, L.-R., and Lee, C.-H.
\newblock A regression approach to speech enhancement based on deep neural
  networks.
\newblock \emph{IEEE/ACM Transactions on Audio, Speech, and Language
  Processing}, 23\penalty0 (1):\penalty0 7--19, 2014.

\bibitem[Yang et~al.(2020)Yang, Qi, Chen, Ma, and Lee]{yang2020characterizing}
Yang, C.-H., Qi, J., Chen, P.-Y., Ma, X., and Lee, C.-H.
\newblock Characterizing speech adversarial examples using self-attention u-net
  enhancement.
\newblock In \emph{ICASSP 2020-2020 IEEE International Conference on Acoustics,
  Speech and Signal Processing (ICASSP)}, pp.\  3107--3111. IEEE, 2020.

\bibitem[Yang et~al.(2021{\natexlab{a}})Yang, Qi, Chen, Chen, Siniscalchi, Ma,
  and Lee]{yang2020decentralizing}
Yang, C.-H.~H., Qi, J., Chen, S. Y.-C., Chen, P.-Y., Siniscalchi, S.~M., Ma,
  X., and Lee, C.-H.
\newblock Decentralizing feature extraction with quantum convolutional neural
  network for automatic speech recognition.
\newblock In \emph{ICASSP 2021-2021 IEEE International Conference on Acoustics,
  Speech and Signal Processing (ICASSP)}, pp.\  6523--6527. IEEE,
  2021{\natexlab{a}}.

\bibitem[Yang et~al.(2021{\natexlab{b}})Yang, Siniscalchi, and
  Lee]{yang2021pate}
Yang, C.-H.~H., Siniscalchi, S.~M., and Lee, C.-H.
\newblock Pate-aae: Incorporating adversarial autoencoder into private
  aggregation of teacher ensembles for spoken command classification.
\newblock \emph{arXiv preprint arXiv:2104.01271}, 2021{\natexlab{b}}.

\bibitem[Ye \& Dai(2018)Ye and Dai]{ye2018novel}
Ye, R. and Dai, Q.
\newblock A novel transfer learning framework for time series forecasting.
\newblock \emph{Knowledge-Based Systems}, 156:\penalty0 74--99, 2018.

\bibitem[Zhang et~al.(2010)Zhang, Zuo, Zhang, and Zhang]{zhang2010time}
Zhang, D., Zuo, W., Zhang, D., and Zhang, H.
\newblock Time series classification using support vector machine with gaussian
  elastic metric kernel.
\newblock In \emph{2010 20th International Conference on Pattern Recognition},
  pp.\  29--32. IEEE, 2010.

\bibitem[Zhou et~al.(2016)Zhou, Khosla, Lapedriza, Oliva, and
  Torralba]{zhou2016learning}
Zhou, B., Khosla, A., Lapedriza, A., Oliva, A., and Torralba, A.
\newblock Learning deep features for discriminative localization.
\newblock In \emph{Proceedings of the IEEE conference on computer vision and
  pattern recognition}, pp.\  2921--2929, 2016.

\end{thebibliography}
\bibliographystyle{icml2021}

\clearpage
\onecolumn
\appendix
\section*{Appendix}

\section{Proof of \textbf{Lemma 1} }
\label{proof_lemma}
For brevity we use $[K]$ to denote the integer set $\{1,2,\ldots,K\}$.
We have
\begin{align}
\bbE_{x \sim \cD,~x' \sim \cD'}\|f(x) - f(x') \|_2 
&\overset{(a)}{=} 
    \bbE_{x \sim \cD,~x' \sim \cD'}\|\eta(z(x)) - \eta(z(x')) \|_2  \\
&\overset{(b)}{=} \int_{x \sim \cD,~x' \sim \cD'} \|\eta(z(x)) - \eta(z(x')) \|_2 \Phi_{\cD,\cD'}(x,x') dx dx' \\
&\overset{(c)}{=} \int_{x \sim \cD,~x' \sim \cD'} \|\eta(z(x)) - \eta(z(x')) \|_2 \Phi_{\cD}(x) \cdot \Phi_{\cD'}(x') dx dx'\\
&\overset{(d)}{\leq} \sqrt{K} \cdot       \int_{x \sim \cD,~x' \sim \cD'} \max_{k \in [K]} |[\eta(z(x))]_k - [\eta(z(x'))]_k| \Phi_{\cD}(x) \cdot \Phi_{\cD'}(x') dx dx' \\
&\overset{(e)}{\leq} 2\sqrt{K} \cdot   \sup_{g: \bbR^K \mapsto \bbR, \|g\|_{\text{Lip}} \leq 1}    \bbE_{x \sim \cD}  [g(z(x))] -  \bbE_{x' \sim \cD'} [ g(z(x'))]\\
&\overset{(f)}{=}  2\sqrt{K} \cdot \cW_1(\mu_z,\mu_z')
\end{align}
(a) follows the neural network model,
(b) follows the definition of expectation, (c) follows the assumption of independent data drawing, and (d) follows that $\|x\|_2 = \sqrt{\sum_{i}^d x_i^2} \leq \sqrt{d \cdot \max_i [x_i^2]}=\sqrt{d} \cdot \max_i |x_i|$, and thus
$\|\eta - \eta'\|_2 \leq \sqrt{K} \cdot  \max_k |[\eta - \eta']_k|$.
(e) holds by setting $k^{+} = \arg \max_{k \in [K]}  [\eta(z(x))]_k - [\eta(z(x'))]_k $ and $k^{-} = \arg \max_{k \in [K]}  [\eta(z(x'))]_k - [\eta(z(x))]_k  $. Then by definition $\max_{k \in [K]} |[\eta(z(x))]_k - [\eta(z(x'))]_k| \leq [\eta(z(x))]_{k^+} - [\eta(z(x'))]_{k^+} + [\eta(z(x'))]_{k^-} - [\eta(z(x))]_{k^-} $. We further make the following three notes: (i) $[\eta(z(x))]_{k^+} - [\eta(z(x'))]_{k^+} \geq 0$ and  $[\eta(z(x))]_{k^-} - [\eta(z(x'))]_{k^-} \geq 0$; (ii) $|a|=\max\{a,-a\}$, and if $a,b \geq 0, \max\{a,b\} \leq a+b$; (iii) There exist at least one $k$ such that $[\eta(x)]_k- [\eta(x)]_k \geq 0.$ One can use proof by contradiction to show (iii) is true. If $[\eta(x)]_k- [\eta(x)]_k < 0$ for every $k$, then summing over $k$ we get a contradiction that $1<1$. Therefore,
\begin{align}
&\int_{x \sim \cD,~x' \sim \cD'} \max_{k \in [K]} |[\eta(z(x))]_k - [\eta(z(x'))]_k| \Phi_{\cD}(x) \cdot \Phi_{\cD'}(x') dx dx' \\
&\leq \int_{x \sim \cD,~x' \sim \cD'} 
\left(
[\eta(z(x))]_{k^+} - [\eta(z(x))]_{k^-} + [\eta(z(x'))]_{k^-} - [\eta(z(x'))]_{k^+}  
\right) 
\Phi_{\cD}(x) \cdot \Phi_{\cD'}(x') dx dx' \\
&= \bbE_{x \sim \cD} [ [[\eta(z(x))]_{k^+} - [\eta(z(x))]_{k^-}] - \bbE_{x' \sim \cD'} [ [\eta(z(x'))]_{k^+} - [\eta(z(x'))]_{k^-}] \\
& \leq 2 \cdot \sup_{g: \bbR^K \mapsto \bbR, \|g\|_{\text{Lip}} \leq 1}    \bbE_{x \sim \cD} [g(z(x))] - \bbE_{x' \sim \cD'} [ g(z(x'))]
\end{align}
where $\|g\|_{\text{Lip}}$ is defined as $\sup_{x,x'} |g(x)-g(x')|/\|x-x'\|_2$, and we use the fact that $[\eta(z)]_k$ is 1-Lipschitz for any $k \in [K]$ \cite{gao2017properties} (so $[\eta]_{k^{+}}-[\eta]_{k^{-}}$ is 2-Lipschitz). Finally,
(f) follows the Kantorovich-Rubinstein theorem \cite{Kantorovich_Rubinstein_thm} of the dual representation of the Wasserstein-1 distance.

\section{Proof of Theorem 1}
\label{proof_thm}
First, we decompose the target risk function as
\begin{align}
\ell_{\cT}(x_t+\delta^*,y_t) &\overset{(a)}{=} 
\ell_{\cS}(x_t+\delta^*,y_s) \\
&\overset{(b)}{=} 
\|f_{\cS}(x_t+\delta^*) - y_s\|_2 \\
&\overset{(c)}{=} 
\|f_{\cS}(x_t+\delta^*) - f_{\cS}(x_s) +  f_{\cS}(x_s) - y_s\|_2 \\
&\overset{(d)}{\leq}  
\underbrace{\|f_{\cS}(x_t+\delta^*) - f_{\cS}(x_s) \|_2}_{A} + \underbrace{\|  f_{\cS}(x_s) - y_s\|_2}_{B} \label{eqn_thm_last}
\end{align}
(a) is based on Assumption 3, (b) is based on the definition of risk function, (c) is by subtracting and adding the same term $f_{\cS}(x_s)$, and (d) is based on the triangle inequality.

Note that by Assumption 1, $\bbE_{\cD_{\cS}} B = \bbE_{\cD_{\cS}} [\ell (x_s,y_s)] = \epsilon_{\cS}$.
Next, we proceed to bound $\bbE_{\cD_{\cS},\cD_{\cT}} A \triangleq \bbE_{x_s\sim\cD_{\cS},x_t\sim \cD_{\cT}} A$. Using Lemma 1,
we have
\begin{align}
\bbE_{\cD_{\cS},\cD_{\cT}} A 
 \leq 2\sqrt{K} \cdot \cW_1(\mu(z_{\cS}(x_t+\delta^*)),\mu(z_{\cS}(x_s)))_{x_t \sim \cD_{\cT},~x_s \sim \cD_{\cS}}  
\end{align}

Finally, take $\bbE_{\cD_{\cS},\cD_{\cT}}$ on both sides of equation (\ref{eqn_thm_last}) completes the proof.

\section{Pre-Trained Model Studies}
\label{sub:sect:c}
We provide advanced studies over different pre-trained acoustic architectures and the associated time series classification performance. In particular, we select the models below, which have attained competitive performance tested on Google Speech Commands version 2 dataset~\cite{warden2018speech} or shown cutting-edge performance (ResNet~\cite{he2016deep}) in the acoustic scene (VGGish~\cite{hershey2017cnn}) and time series classification (TCN~\cite{lea2016temporal}). These models will be compared with V2S$_a$ (recurrent Attention~\cite{de2018neural}) and V2S$_u$ (V2S$_a$ enhanced by U-Net~\cite{yang2020decentralizing}) used in the main paper.   

\textbf{ResNet:} Deep residual network~\cite{he2016deep} (ResNet) is a popular deep architecture to resolve the gradient vanish issues by passing latent features with a residual connection, and it has been widely used in acoustic modeling tasks. We select a 34-layer ResNet model training from the scratch for V2S (denoted as V2S$_r$), which follows the identical parameter settings in~\cite{hu2020device, hu2021two} for reproducible studies. 

\textbf{VGGish:} VGGish~\cite{hershey2017cnn} is a deep and wide neural network architecture with multi-channel convolution layers, which has been proposed for speech and acoustic modeling. VGGish is also well-known for the large-scale acoustic embedding studies with Audio-Set~\cite{gemmeke2017audio} from 2 million Youtube audios. We use the same architecture and train two models: (i) training from scratch (denoted as V2S$_v$) and  (ii) selecting an Audio-Set pretrained VGGish and then fine-tuning (denoted as V2S$_{p}$) on the Google Speech commands dataset for V2S.  

\textbf{Temporal Convolution Network (TCN):} TCN~\cite{lea2016temporal} is an efficient architecture using temporal convolution with causal kernel for sequence classification tasks. We select the TCN architecture and train it from scratch as a baseline for V2S (denoted as V2S$_t$).

\textbf{OpenL3:} OpenL3~\cite{cramer2019look} is a much recent embedding method with a deep fusion layer for acoustic modeling. We use pretrained OpenL3 embeddings and a dense layer for classification as another baseline for V2S (denoted as V2S$_o$). 
\subsection{V2S Performance and Sliced Wasserstein Distance}
Table~\ref{tab:sub:v2s:more} shows different neural architectures for acoustic modeling to be used with the proposed V2S method, where acoustic models are pre-trained with the Google Speech Commands dataset~\cite{warden2018speech} version two with 32 commands (denoted as GSCv2.) From the first three rows of Table~\ref{tab:sub:v2s:more}, we observe the recurrent attention models and TCN perform better in mean prediction accuracy, validation loss of the source task. For the target task (same as Table~\ref{tab:ucr:1}), recurrent attention models (V2S$_a$ and V2S$_u$) attain the best performance. 

\begin{table}[ht!]
\centering
\caption{V2S ablation studies with different pre-trained acoustic models.}
\label{tab:sub:v2s:more}
\begin{tabular}{|l|l|l|l|l|l|l|l|}
\hline
Model & V2S$_a$           & V2S$_u$  & V2S$_r$  & V2S$_v$  & V2S$_p$  & V2S$_t$           & V2S$_o$  \\ \hline \hline
Parameters ($\downarrow$) & \textbf{0.2M}  & 0.3M  & 1M    & 62M   & 62M   & 1M             & 4.7M  \\ \hline
Source Acc. ($\uparrow$) & 96.90          & 96.92 & 96.40 & 95.40 & 95.19 & \textbf{96.93} & 92.34 \\ \hline
Source Loss ($\downarrow$) & \textbf{0.1709} & 0.1734 & 0.1786 & 0.1947 & 0.1983 & 0.1756          & 0.2145 \\ \hline \hline
Mean SWD ($\downarrow$) & \textbf{1.829} & 1.873 & 1.894 & 4.716 & 4.956 & 1.901          & 5.305 \\ \hline
Mean Target Acc. ($\uparrow$) & \textbf{89.37} & 87.36 & 87.01 & 67.00 & 62.83 & 86.21         & 60.34 \\ \hline
Target MPCE ($\downarrow$) & \textbf{2.04}  & 2.13  & 2.26  & 34.1  & 39.1  & 2.36           & 41.34 \\ \hline
\end{tabular}
\end{table}

Both VGGish based architectures (V2S$_v$ and V2S$_p$) show degraded performance on the target tasks prediction, which can be explained by the recent findings~\cite{kloberdanz2020reprogramming} on the degraded performance of using VGG~\cite{simonyan2014very} and MobileNet-V2~\cite{sandler2018mobilenetv2} based ``wide'' and deep convolutional neural architectures for reprogramming visual models.  These findings could be also explained in the sense that the wide neural architectures fail to adapt the source domain distribution according to the sliced Wasserstein distance (SWD) results (fourth row in Table~\ref{tab:sub:v2s:more}), as  indicated by Theorem~1. We observe that using the pretrained models associated with higher source accuracy does not always guarantee higher target accuracy (e.g, V2S$_t$ and V2S$_u$)

\section{Additional Ablation Studies}
\label{sub:sect:d}
Based on the discussion in Section \ref{sub:sect:c}, we further select three efficient V2S models with different model capacity (0.2M/1M/4.7M), including V2S$_a$, V2S$_r$, and V2S$_t$, to study the mean target accuracy in different training settings and to provide some insights into effective design of V2S models.

\subsection{Pretrained Models from Different Dataset}
In the previous model reprogramming studies~\cite{tsai2020transfer, elsayed2018adversarial}, little has been discussed regarding the effectiveness of using different datasets to train the pretrained models. We study three other public acoustic classification benchmark datasets (source tasks), (1) TAU Urban Acoustic Scenes 2020 Mobile~\cite{heittola2020acoustic} (denoted as TAU-UAC), from the annual IEEE AASP Challenge on Detection and Classification of Acoustic Scenes and Events (DCASE), (2) AudioSet using in~\cite{hershey2017cnn}, and (3) ESC-50~\cite{piczak2015dataset}, a dataset for environmental sound classification, for providing a preliminary V2S study.
We extract acoustic features by Mel-spectrogram using Kapre audio layer following the same resolution and frequency setup in ~\cite{yang2020characterizing, yang2021pate} for a fair and reproducible comparison. 

As shown in Table~\ref{tab:sub:pre:more}, the V2S models pretrained from GSCv2 show a higher mean prediction accuracy and lower SWD than the other models, which could be due to the shorter sample length (less than one second) of its source acoustic inputs.   

\begin{table}[ht!]
\centering
\caption{V2S performance (mean prediction accuracy) on time series classification (same as Table \ref{tab:ucr:1}) with different pretrained neural acoustic models and the mean sliced Wasserstein distance for each dataset.}
\label{tab:sub:pre:more}
\begin{tabular}{|l|l|l|l|l|}
\hline
Pretrained Dataset & GSCv2          & TAU   & AudioSet & ESC   \\ \hline \hline
$\#$ Training Samples  & 105k & 13.9k & 2.08M    & 2k \\ \hline
$\#$ Output Classes & 35 & 10    & 527 & 50 \\ \hline Audio length per sample clips & 1 sec. & 10 sec. & 10 sec.    & 5 sec. \\ \hline \hline Mean Target Acc. w/ V2S$_a$ ($\uparrow$)              & \textbf{89.37} & 82.61 & 80.68    & 84.48 \\ \hline
Mean Target Acc. w/ V2S$_r$ ($\uparrow$)              & 87.36          & 83.57 & 79.96    & 83.05 \\ \hline
Mean Target Acc. w/ V2S$_t$ ($\uparrow$)              & 86.31          & 80.1  & 81.81    & 84.58 \\ \hline \hline
Mean SWD per dataset ($\downarrow$)          & \textbf{1.882} & 2.267 & 2.481    & 2.162 \\ \hline
\end{tabular}
\end{table}

\subsection{Different V2S Mapping Settings}
\label{sub:sec:v2s}
In \cite{tsai2020transfer}, frequency mapping techniques show improved performance for black-box (e.g., zeroth order gradient estimation~\cite{chen2017zoo,liu2020primer} based) adversarial reprogramming models for image classification. We also follow the setup in \cite{tsai2020transfer} to compare the many-to-one frequency mapping and the many-to-one random mapping for time series classification. However, the frequency mapping based V2S results show equal or slightly worse (-0.013$\%$) mean prediction accuracy and WSD (+0.0028) performance with 100 runs, which may be owing to the differences of the dimensions and scales between the tasks of image and time series reprogramming. 

\subsection{More tSNE Visualization}
We provide more tSNE visualization over different test datasets to better understand the embedding results of V2S models discussed in Section \ref{subsec_visual}. In Figure~\ref{fig:sup:more:tsne} (a) to (e), the reprogrammed representations (rightmost side) show better disentangled results in both 2D and 3D tSNE plots. 

\subsection{Hardware Setup and Energy Cost Discussion}
We use Nvidia GPUs (2080-Ti and V100) for our experiments with Compute Unified Device Architecture (CUDA) version 10.1. To conduct the results shown in Table~\ref{tab:ucr:1}, it takes around 40 min to run 100 epochs (maximum) with a batch size 32 for each time series prediction dataset considering the hyper-parameters tuning (e.g., dropout rate) described in Section \ref{subsec_implement} of the main paper. In total, the experiments presented (30 datasets and its ablation studies) in this paper took around 120 computing hours with a 300W power supplier. As another advantage, the V2S reprogramming techniques freeze pretrained neural models and only used a reprogramming layer for training new tasks. The proposed method could potentially recycle well-trained models for an additional task to alleviate extra energy costs toward deploying responsible ML systems.

\begin{figure}[ht!]
	\begin{subfigure}{0.46\textwidth} %
	    \centering
	    \includegraphics[width=\textwidth]{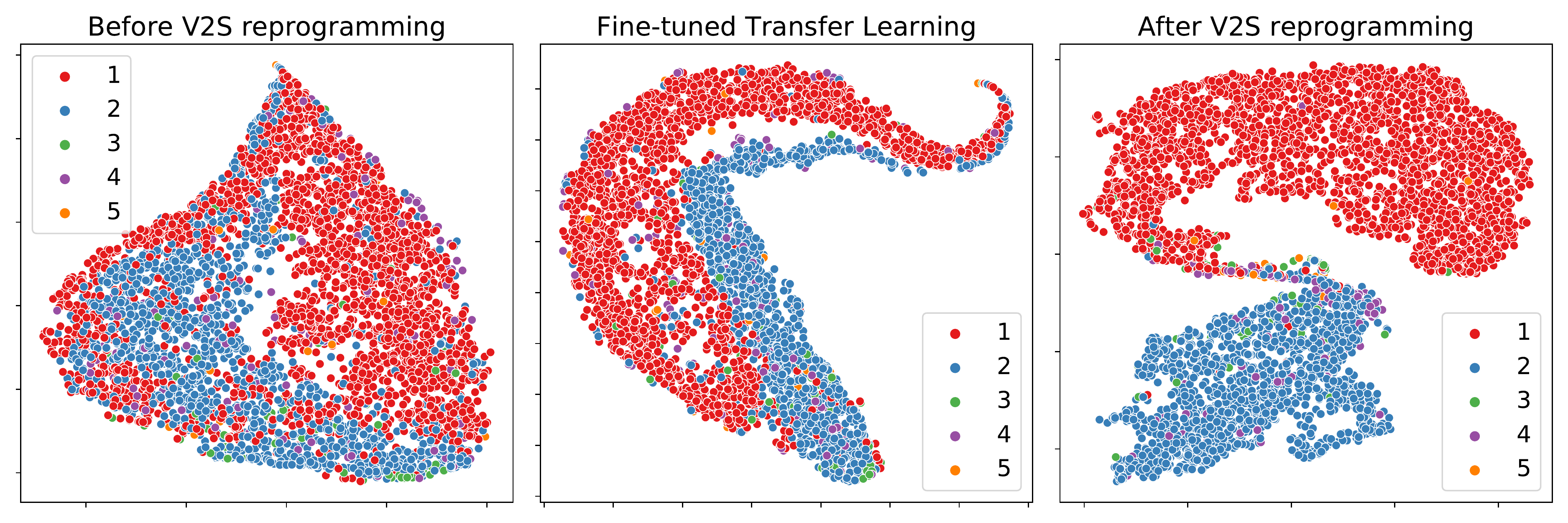}
	    \caption{Task: ECG 5000 with 2D tSNE} %
	\end{subfigure}
	\quad
	\begin{subfigure}{0.46\textwidth} %
	    \centering
		\includegraphics[width=\textwidth]{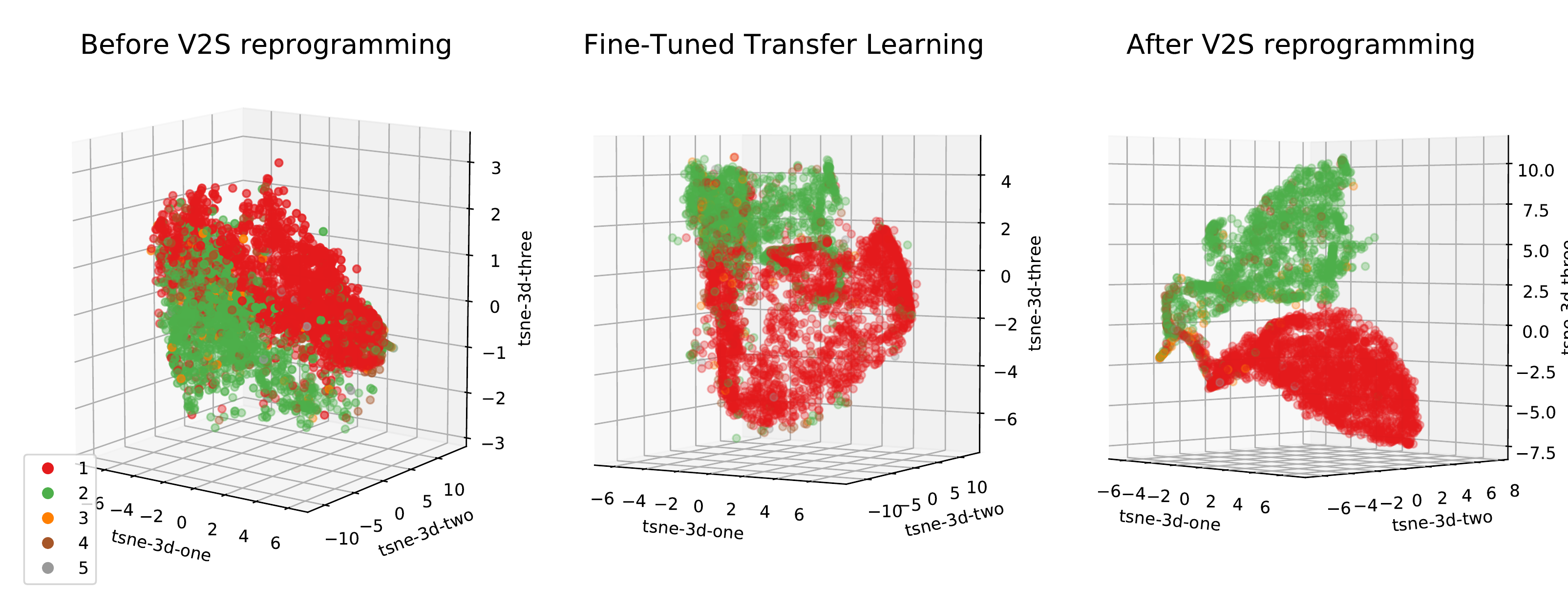}
		\caption{Task: ECG 5000 with 3D tSNE } %
	\end{subfigure}
	\quad
	\begin{subfigure}{0.46\textwidth} %
	    \centering
		\includegraphics[width=\textwidth]{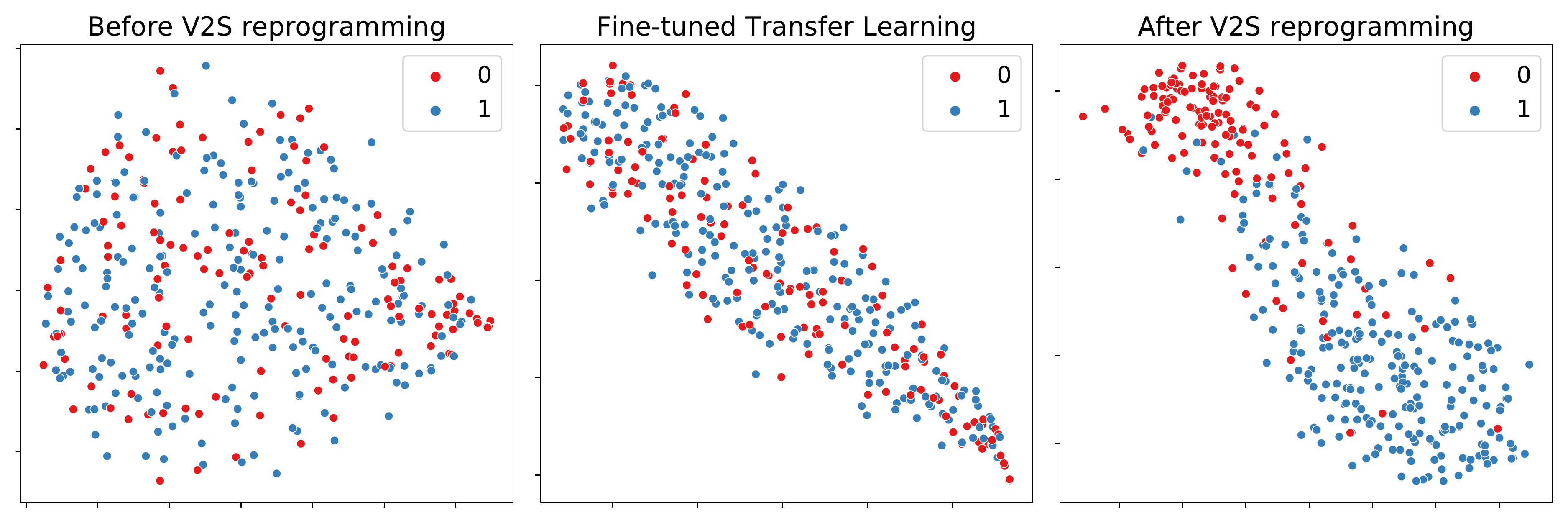}
		\caption{Task: HandOutlines with 2D tSNE.} %
	\end{subfigure}
	\quad %
	\begin{subfigure}{0.46\textwidth} %
	    \centering
		\includegraphics[width=\textwidth]{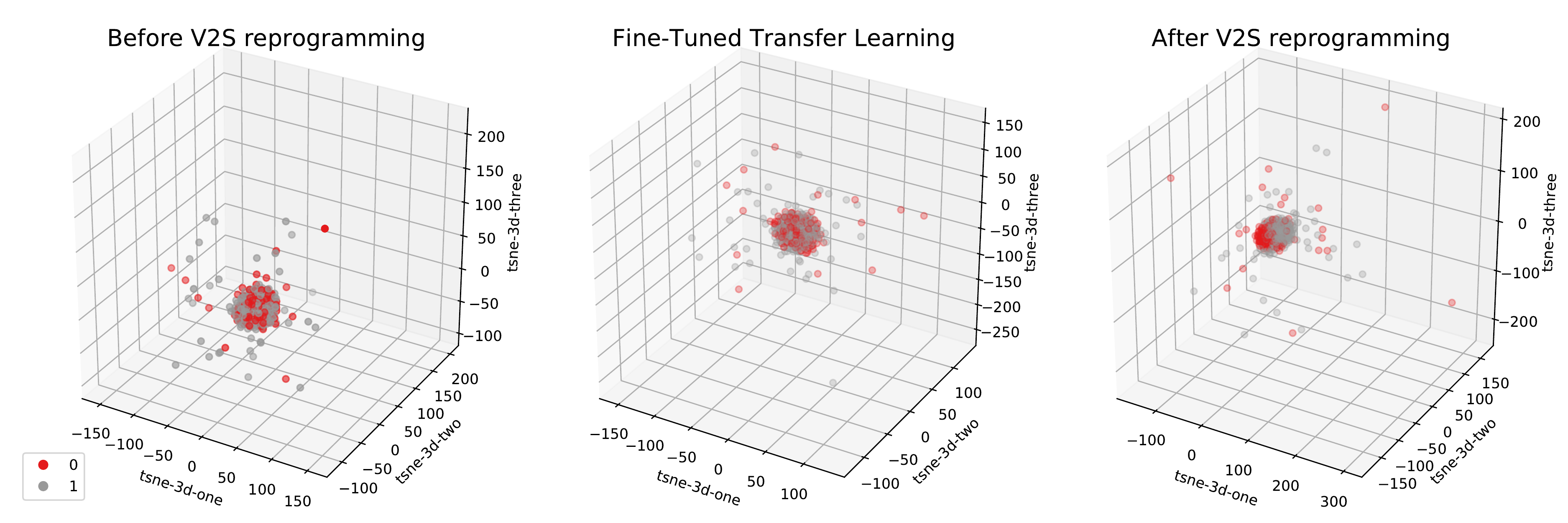}
		\caption{Task: HandOutlines with 3D tSNE.} %
		\end{subfigure}
	 \begin{subfigure}{0.46\textwidth} %
	    \centering
	    \includegraphics[width=\textwidth]{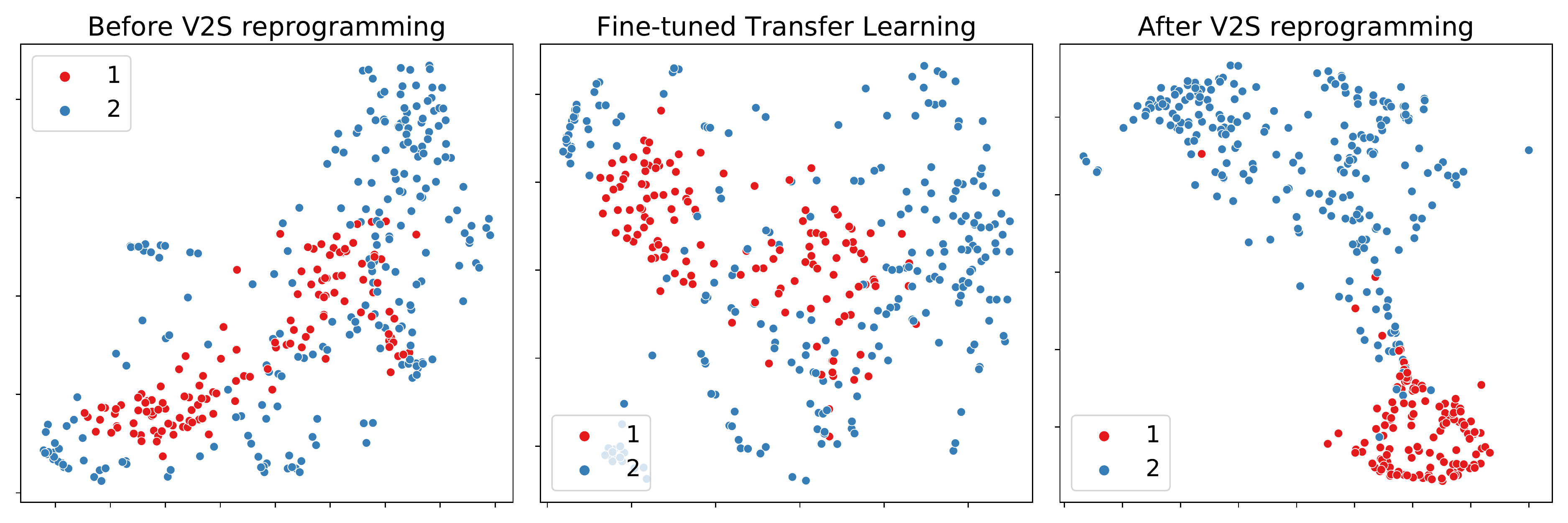}
	    \caption{Task: Strawberry 2D tSNE.} %
	\end{subfigure}
	\quad\quad\quad\quad
	\begin{subfigure}{0.46\textwidth} %
	    \centering
		\includegraphics[width=\textwidth]{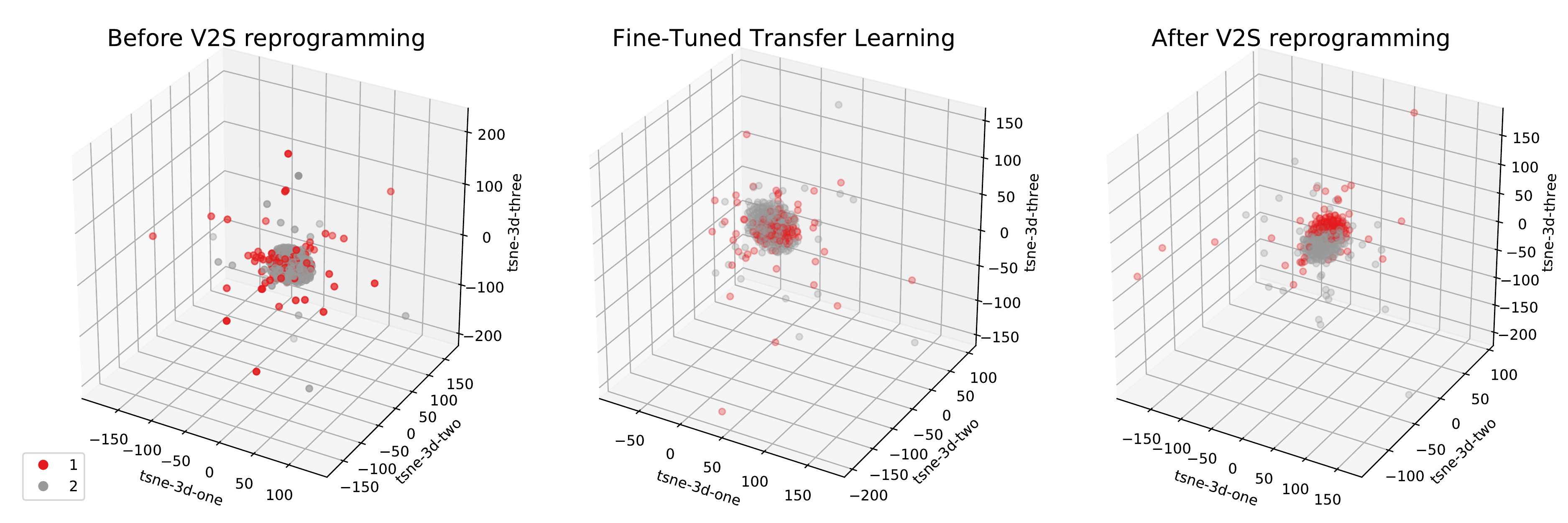}
		\caption{Task: Strawberry 3D tSNE } %
	\end{subfigure}
	\caption{More tSNE visualization. Numbers in the legend are class label indices. } %
	\label{fig:sup:more:tsne}

\end{figure}

\end{document}